\documentclass[10pt,journal,compsoc]{IEEEtran}

\ifCLASSOPTIONcompsoc
  \usepackage[nocompress]{cite}
\else
  \usepackage{cite}
\fi

\usepackage{amsmath,amsfonts}
\usepackage{algorithmic}
\usepackage{array}
\usepackage[caption=false,font=normalsize,labelfont=sf,textfont=sf]{subfig}
\usepackage{textcomp}
\usepackage{stfloats}
\usepackage{url}
\usepackage{verbatim}
\usepackage{graphicx}
\def\BibTeX{{\rm B\kern-.05em{\sc i\kern-.025em b}\kern-.08em
		T\kern-.1667em\lower.7ex\hbox{E}\kern-.125emX}}
\usepackage{balance}

\usepackage{amsmath,bm}
\usepackage{amsfonts,amssymb}
\usepackage{booktabs}
\usepackage{graphicx} 
\usepackage{float}
\usepackage{bm}
\usepackage{color}
\usepackage{multirow}
\newcommand{\vts}{\mathrm{T}}
\newcommand{\bv}[1]{{\bm {#1}}}
\newcommand{\ba}[1]{{\bf {#1}}}
\newcommand{\beq}{\begin{equation}}
	\newcommand{\eeq}{\end{equation}}
\newcommand{\bmat}{\begin{bmatrix}}
	\newcommand{\emat}{\end{bmatrix}}
\newcommand{\Eq}[1]{(\ref{#1})}
\newcommand{\Fig}[1]{Fig. \ref{#1}}
\newcommand{\Tab}[1]{Tab. \ref{#1}}

\newcommand{\Sec}[1]{Section \ref{#1}}

\newcommand{\anno}[1]{{\color{black}{#1}}}

\ifCLASSINFOpdf
\else
\fi
\begin{document}
%
\title{TextSLAM: Visual SLAM with Semantic Planar Text Features}		

%
%
%

\author{Boying Li, 
		Danping Zou*, 
		Yuan Huang, 
		Xinghan Niu, 
		Ling Pei, 
		and Wenxian Yu 
	
        
        
        
\IEEEcompsocitemizethanks{\IEEEcompsocthanksitem All the authors are with Shanghai Key Laboratory of Navigation and Location-Based Services, Shanghai Jiao Tong University.\protect\\
	Corresponding author*: Danping Zou (dpzou@sjtu.edu.cn).
}

}

%
%

\markboth{Journal of \LaTeX\ Class Files,~Vol.~14, No.~8, August~2015}%
{Shell \MakeLowercase{\textit{et al.}}: Bare Advanced Demo of IEEEtran.cls for IEEE Computer Society Journals}
%



\IEEEtitleabstractindextext{%
\begin{abstract}
		We propose a novel visual SLAM method that integrates text objects tightly by treating them as semantic features via fully exploring their geometric and semantic prior. The text object is modeled as a texture-rich planar patch whose semantic meaning is extracted and updated on the fly for better data association. With the full exploration of locally planar characteristics and semantic meaning of text objects, the SLAM system becomes more accurate and robust even under challenging conditions such as image blurring, large viewpoint changes,  and significant illumination variations (day and night). \anno{We tested our method in various scenes with the ground truth data. The results show that integrating texture features leads to a more superior SLAM system that can match images across day and night.} The reconstructed semantic 3D text map could be useful for navigation and scene understanding in robotic and mixed reality applications.	
		\anno{Our project page: \url{https://github.com/SJTU-ViSYS/TextSLAM}}.
		

\end{abstract}

}

\maketitle

\IEEEdisplaynontitleabstractindextext

%
\IEEEpeerreviewmaketitle

\ifCLASSOPTIONcompsoc
\IEEEraisesectionheading{\section{Introduction}\label{sec:introduction}}
\else
\section{Introduction}
\label{sec:introduction}
\fi
	\IEEEPARstart{V}{isual} SLAM is an important technique of ego-motion estimation and scene perception, which has been widely used in navigation for drones \cite{heng2014autonomous}, ground vehicles, self-driving cars \cite{lategahn2011visual}, or other applications such as  Augmented and Virtual Reality (AR and VR) \cite{chekhlov2007ninja}. 
	The typical visual SLAM algorithm extracts point features \cite{mur2015orb,davison2003real} from images for pose estimation and mapping.
	Recent methods \cite{forster2014svo,engel2018direct} even directly operate on pixels.
	However, it is well known that incorporating high-level features like lines \cite{zhou2015structslam}, surfaces \cite{trevor2012planar} or even semantic objects {\cite{lianos2018vso,mccormac2017semanticfusion}} in the visual SLAM system will lead to better performance.

	\anno{One common type of high-level feature is text objects.}  
	Scene texts play a key role in identifying locations with various forms such as road marks \cite{ranganathan2013light,radwan2016you}, building or object signs \cite{li2020textslam, hong2019textplace}, room names \cite{rong2016guided,li2019vision, hong2019textplace}, and other textual captions \cite{wang2015bridging,rong2016guided, li2019vision}. They help us to recognize landmarks, navigate in complex environments, and guide us to the destination. 
	Detection and recognition of scene texts from images have been developing fast \cite{zhang2019feasible, zhou2017east, lyu2018mask, wang2020all, liao2020real, he2021most, zhu2016scene, yin2016text} because of the boom of deep neural networks and the emergence of huge text datasets such as COCO-Text \cite{veit2016coco}, DOST \cite{iwamura2016downtown}, and ICDAR \cite{chng2019icdar2019}. As extracting scene texts from images becomes easy nowadays, one question raises whether texts can be integrated into a visual SLAM system to both yield better performance and generate high-quality 3D semantic text maps that could be useful for robot navigation and scene understanding, as well as augmented reality and human-computer interaction.
	
	\begin{figure}[t]		
		\centering  
		\includegraphics[width=0.49\textwidth]{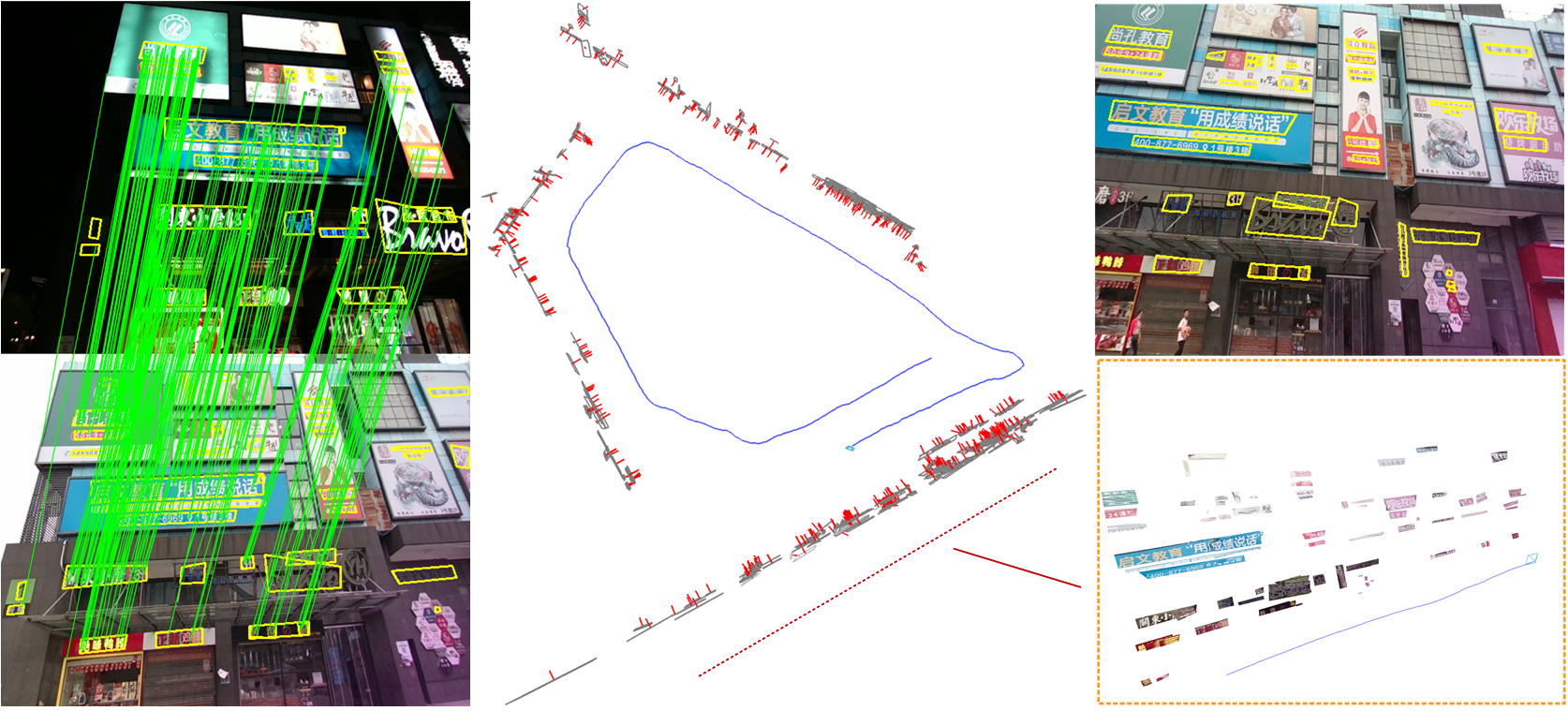}\\ 	
		\caption{\anno{TextSLAM can produce 3D text maps and match text objects correctly despite significant illumination changes.} \textbf{Left column}: semantically matched text objects and text-guided point correspondences (green lines)  between a night query image and a day image in the day-and-night test. The detected texts are shown in yellow rectangles. \textbf{Middle column}: \anno{3D text maps and camera trajectory in the bird-eye view.} The text objects are illustrated in gray boxes and their normal directions are shown in red. \textbf{Right column}: \anno{tracked texts in the image} (in yellow rectangles) and the zoomed-in view of the 3D text map.}
		\label{fig:textEx}  
	\end{figure}
	
	Texts spotted in our daily life are mostly planar regions, at least for a single word or character if not the whole sentence. The rich texture and planar property of a text entity make the text object a good feature for tracking and localization.
	More importantly, the semantic messages that a text object delivers are invariant to appearance changes, hence text objects are also reliable features for matching even when the illumination or viewpoint changes significantly. Those characteristics of scene texts are certainly good for SLAM, while the key issue is how to integrate them into a visual SLAM system.
	
	There are several attempts towards coupling SLAM with text features. 
	A navigation system with human-computer interaction \cite{rong2016guided, li2019vision} for blind people, assisted with text entities, is built upon the visual-inertial SLAM system shipped on Google Tango tablet.
	Similarly, Wang et al. proposed a method to extract text features \cite{wang2015bridging}, which are then used for fusion with Tango's SLAM outputs to recognize revisited places. 
	The aforementioned works have shown the great potential of using text features with existing SLAM systems. However, they treat the SLAM system as a black box, which is unable to fully take advantage of the characteristics of scene texts that should be beneficial to SLAM.
	
	In this paper, we present a novel SLAM system tightly coupled with semantic text features as shown in \Fig{fig:textEx}. Specifically, we integrate the text features into the SLAM pipeline by fully exploiting the favorable characteristics of scene texts. Geometrically, text features are treated as texture-rich planar patches used for camera pose estimation and back-end optimization to yield more accurate and robust estimation. Semantically, the meaning of those scene texts, invariant to appearance changes, are utilized for reliable place recognition and feature matching across scenes with large illumination or viewpoint changes.  
	\anno{For the lack of SLAM benchmarks with rich texts, we collected a text-orientated dataset both indoor and outdoor with the ground truth carefully acquired for evaluation. We compare our SLAM system with state-of-the-art approaches.} 
	The results show that by tightly coupling with text objects, the SLAM system becomes more accurate and robust, and even works well under challenging conditions such as serious illumination changes (day and night), viewpoint changes, and occlusions, where existing SLAM systems usually fail.
	\anno{We also compared our text-based method with the state-of-the-art visual localization methods for loop closing. The results show that our text-based method outperforms those methods in text-rich environments with a much lower computational cost.}
	
	The technical contributions of our work include:
	
	1) A novel visual SLAM framework that integrates text features in front-end pose tracking, back-end optimization, loop closing, and mapping. To our best knowledge, this is the first work that tightly integrates scene texts into the pipeline of visual SLAM.
	\anno{We also contribute a dataset in various text-rich environments for evaluation.}

	2) We present both geometric and semantic representations of text features, as well as their observation and data association models within the SLAM pipeline.
	
	3) A novel loop closing technique relying on the semantic meaning of text features. With the help of semantic information, reliable loop closing can be achieved even in challenging scenarios, including serious illumination changes, occlusion, and drastically varying viewpoints.
	
	
	
	\anno{This paper is extended from our previous work \cite{li2020textslam}. The major extension is incorporating the \textbf{semantic information} into the SLAM pipeline, especially for the semantic data association and loop closure, as well as additional experiments and analysis.
	Specifically, the extensions include a novel semantic representation of text objects together with its update scheme (Section 3.2), using the semantic information of text objects for loop closure (Section 4.5), and several improvements of the SLAM system (Section 4) such as text object culling, text selection in pose estimation, and feature sampling in the coarse-to-fine optimization. 
	In addition, we present a challenging text-orientated dataset with ground truth. Additional tests in indoor, outdoor and day-night switching are also presented in Section 5.}

 \begin{table*}[]
 	\caption{A list of existing text-aided navigation approaches}
 	\scriptsize
 	\setlength{\tabcolsep}{1.9mm}
 	\begin{tabular}{lllllll}
 		\hline
 		{\bf Method}  & {\bf Text objects} & {\bf Text extraction} & {\bf Map} & {\bf Task} & {\bf Scene} \\
 		\hline
 		Tomono et al. 2000 \cite{tomono2000mobile}  & room nameplates & heuristic & office map with a corridor and doors & Robot navigation &  indoor \\
 		Mata et al. 2001 \cite{mata2001visual}       & room nameplates &  heuristic & office map with landmark annotation & Robot navigation &  indoor \\
 		Case et al. 2011 \cite{case2011autonomous}   & room nameplates & heuristic  & laser grid-based map with text annotation & Robot navigation & indoor \\ 
 		Ranganathan et al. 2013 \cite{ranganathan2013light} & road marks & heuristic & road surface marks map & localization &  outdoor \\  
 		Wang et al. 2015 \cite{wang2015bridging}     & artificial tags & heuristic & landmarker map &  loop closing & indoor \\
 		\anno{Wang et al. 2015 \cite{wang2015lost}}		& \anno{store signs} & \anno{heuristic} & \anno{floorplan with text annotation} &  \anno{localization} & \anno{indoor} \\
 		Radwan et al. 2016 \cite{radwan2016you}        & store signs & heuristic & geo-tagged street-level map with text annotation &  localization & outdoor \\   
 		Hong et al. 2019 \cite{hong2019textplace}    & \anno{street\&store signs}  & deep learning  & \anno{2D imagery map} &  localization  &  indoor\&outdoor \\	
 		Li et al. 2019 \cite{li2019vision}         & room nameplates &  heuristic & CAD model map with semantic layers & human navigation & indoor \\ 
 		\hline
 		{\bf TextSLAM (Ours)} & all scene texts  & deep learning & 3D text map & tightly coupled SLAM & indoor\&outdoor \\
 		\hline  
 	\end{tabular}
 	\label{tab:relatework_text_aided_nav}
 \end{table*}

\section{Related work}
	\subsection{Planar features}
	Most scene texts can be treated as texture-rich planar features.
	Planar features have been studied in the visual SLAM community since the early stage. In early works \cite{chekhlov2007ninja}\cite{gee2007discovering}\cite{gee2008discovering}, the planes in the scene were detected by RANSAC \cite{yang2010plane} among estimated 3D points and employed as novel features to replace points in the state. Since much fewer parameters are required to represent the map using planes, it reduces the computational cost significantly\cite{davison2019futuremapping}. 
	These works show that planar features improve both accuracy and robustness of a visual SLAM system.
	Existing methods require 3D information to discover the planar features, usually using a RGB-D camera \cite{sturm2012benchmark}\cite{ma2016cpa}\cite{kim2018linear}. However, this becomes difficult using only image sensors. An interesting idea \cite{molton2004locally} is to assume the region surrounding the detected feature point as a locally planar surface. The assumption seldom holds in realistic scenes, as feature points might be extracted from anywhere in the scene.
	By contrast, texts in realistic scenes are mostly located on planar surfaces. Unlike general planar features that usually require depth for  detection \cite{sturm2012benchmark}\cite{ma2016cpa}\cite{kim2018linear}, scene texts can be easily extracted by off-the-shelf text detectors \cite{zhang2019feasible, zhou2017east}. 
	
	\subsection{Visual SLAM with semantics}
Integrating semantic information into visual SLAM systems has been receiving increasing interest in recent years \cite{sualeh2019simultaneous,davison2018futuremapping,davison2019futuremapping,rosinol2020kimera,chang2021kimera}. One approach is to directly fuse the 2D semantic labels with the dense 3D map from RGB-D SLAM \cite{mccormac2017semanticfusion, runz2018maskfusion, grinvald2019volumetric}, or a dense visual SLAM \cite{zhi2019scenecode}.
	Another approach is to take semantic objects as high-level features within the SLAM pipeline \cite{galvez2016real}\cite{salas2013slam++}, which requires pre-scanned 3D models to precisely fit the observation on the image. 
	\anno{Though recent methods \cite{mccormac2018fusion++, laidlow2022simultaneous, mazur2022feature, xu2022learning} build the 3D representation of objects online with a depth camera, it is still difficult to be generalized to unseen objects with only video cameras. }	
	Other methods seek to use 3D bounding boxes \cite{yang2019cubeslam}\cite{yang2019monocular}\cite{dong2017visual} or quadrics \cite{nicholson2018quadricslam} to represent objects, but such kind of approximation suffers from loss of accuracy.
	In this paper, we focus on the particular semantic object, i.e., scene texts.
	Unlike generic objects, text objects, such as road signs, shop signs, room numbers, and commercial boards, are texture-rich planar features and contain rich semantic information about environments. Those characteristics are more favorable to visual SLAM  than those of general objects.

    \subsection{Text-aided navigation}
Scene Texts such as room numbers, road marks, route or traffic signs, and shop signs are naturally good visual landmarks to assist navigation. 
\anno{We summarize existing works on text-aided navigation in \Tab{tab:relatework_text_aided_nav}.}
In the early works \cite{tomono2000mobile} \cite{mata2001visual},  indoor text labels such as room numbers or name tags were used as guidance for a robot to navigate in the lab environments. However, text-aided navigation was still in its infancy at that time, as the technique of detection and recognition of scene texts was still under early development \cite{letourneau2004autonomous,liu2005edge}.
Ranganathan et al. \cite{ranganathan2013light} integrated standard road marks into the pre-built GPS+IMU+camera map to estimate the vehicle's ego-motion for autonomous driving.
With the prior knowledge of a comprehensive geo-tagged street-level map (e.g. GoogleMaps or OpenStreetMap) and the compass information, Radwan et al. \cite{radwan2016you} extracted text information from the street signs to assist pose estimation in a 2D map.
\anno{Similarly, Wang et al. \cite{wang2015lost} used the shop names for localization by taking the building's floor plan as a prior under the assumption of Manhattan world.}
\anno{Following the idea of the aforementioned works, Hong et al.  \cite{hong2019textplace} applied neural networks to extract street and store names together with billboards in the wild for place recognition.}
Those works have shown the advantage of leveraging text objects in dealing with illumination and viewpoint changes for localization. It is natural to consider integrating scene texts to a SLAM system to gain better performance as well as to offer a new way for human-computer interaction.

Wang et al. \cite{wang2015bridging} proposed a spatial-level feature named 'junction' for the text extraction, and then used the text objects to improve loop closing performance based on Google Tango's SLAM outputs. Case et al. \cite{case2011autonomous} annotated the text labels on the existing map generated by a laser SLAM system to help robots recognize named locations and understand humans' free-text queries. 
Rong et al. \cite{rong2016guided} presented an assistive blind navigation system with a text spotting function based on the Tango's SLAM system. Similarly, built on the SLAM system of Tango, a mobile solution \cite{li2019vision} of assistive navigation system combined various sources, such as text recognition results and speech-audio interaction, for blind and visually impaired people to travel indoors independently.
Existing text-aided navigation systems integrate text objects loosely by regarding the SLAM system as a black box. 
By contrast, our proposed method integrates the text objects tightly into the SLAM system to facilitate both camera tracking and semantic mapping.
Moreover, the semantic information from the established map is used for loop closing and camera localization to achieve good performance even under challenging conditions.

\section{Semantic text features}

	A semantic text feature is represented as a planar and bounded patch with its unique semantic meaning. 
	We describe the geometric model (including the parameterization and observation models) of a text object, as well as the way to represent and update the semantic information in the following sections.
\subsection{Geometric model}
\subsubsection{Parameterization}
	\begin{figure}[!tbp]		
		\centering  
		\includegraphics[width=0.38\textwidth]{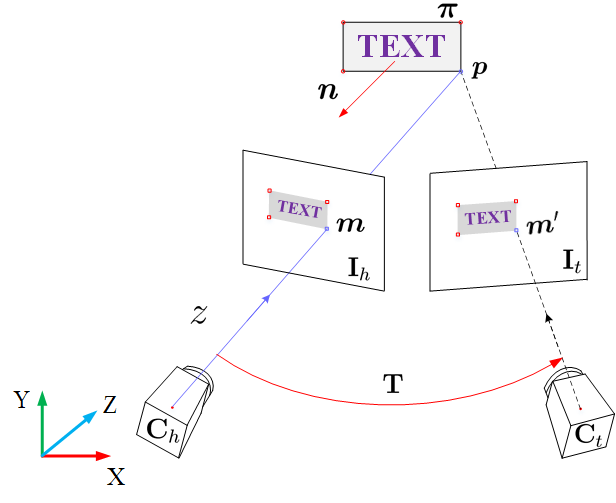}\\	
		\caption{A text object is compactly parameterized by $\bv{\theta}$. The inverse depth $\rho$ of a text point $\bv{p}$ can be computed by $\rho = 1/z = \bv{\theta}^\vts \tilde{\bv{m}}$ and its projection onto the target view $C_t$ is a homography transform with respect to the relative pose $\ba{T}$ between the two views.} 
		\label{fig:textParam}  
	\end{figure}
	
	Text objects are regarded as planar and bounded patches. Each text patch (enclosed by a bounding box) is anchored to a camera frame, named as the \emph{host frame}, which is the first frame where the text object appears as shown in \Fig{fig:textParam}.
	Within the host frame, the plane where the text patch lies is given by
	$ \bv{n}^\vts\bv{p}+d = 0,$	where $\bv{n}=(n_1,n_2,n_3)^\vts\in\mathbb{R}^{3}$ is the normal of the plane and $d \in \mathbb{R}$ is related to the distance from the plane to the origin of the host frame; $\bv{p}\in\mathbb{R}^{3}$ represents the 3D point on the plane.
	
	A straightforward parameterization of a text plane could be directly using the four parameters $(n_1,n_2,n_3,d)$ of the plane equation. However, this is an over-parameterization that leads to rank deficiency in the nonlinear least-squares optimization. We hence adopt a compact parameterization that contains only three parameters.
	\beq
	\bv{\theta}=(\theta_1,\theta_2,\theta_3)^\vts = -\bv{n}/d.
	\eeq  We'll show that this parameterization is closely related to the inverse depth of the 3D point on the text plane.

	Within the host frame, each 3D point $\bv{p} \in \mathbb{R}^{3}$ observed on the image can be represented by its normalized image coordinates $\bv{m}=(u,v)^\vts$ and its inverse depth $\rho = 1/z$. The 3D coordinates of this point are computed as $\bv{p} = (uz,vz,z)^\vts = z\tilde{\bv{m}}$, where $\tilde{\bv{m}}$ represents the homogeneous coordinates of $\bv{m}$. If the 3D point locates on the text plane, we have $z \cdot\bv{n}^\vts\tilde{\bv{m}}+d=0$.	
	The inverse depth $\rho$ of this 3D point is then computed as
	\beq
	\rho = 1/z = - \bv{n}^\vts/d\,\,\tilde{\bv{m}} = \bv{\theta}^\vts 	
	\tilde{\bv{m}}.
	\label{eq:invdepth_01}
	\eeq
	That is, we can use a simple dot product to quickly infer the inverse depth of a text point from its 2D coordinates, given the text parameters $\bv{\theta}$.
	
	On the other hand, if we have at least three points on the text patch (for example, three corners of the bounding box), with their inverse depth value, we can immediately obtain the text parameters by solving
	\beq
	\bmat
	\tilde{\bv{m}}^\vts_1\\
	\vdots\\
	\tilde{\bv{m}}^\vts_n
	\emat \bv{\theta} = 
	\bmat
	\rho_1\\
	\vdots\\
	\rho_n
	\emat, n \ge 3.
	\label{eq:param_from_invdepth}
	\eeq
	This allows us to quickly initialize the text parameters from the depth value of three corners of the text bounding box. 
	
	Properties such as the boundary of a text object are kept in our system. Those properties are acquired from a text detector as we'll describe later. 
\subsubsection{Observation}	

	\anno{To update the parameters of a text object, an observation model should be defined. Here we choose an observation model that measures the difference} between the detected text object and the projection of the estimated 3D text object on the image. In the first step, we need to project the 3D text object from the host frame onto the target image plane.

	Let $\ba{T}_{h}, \ba{T}_{t}\in SE(3)$ represent the transformations of the host frame and the target frame with respect to the world frame. The relative pose  between the host frame and the target frame is computed as $\mathbf{T} =\mathbf{T}_{t}^{-1} \mathbf{T}_{h}$. We let $\ba{R},\bv{t}$ be the rotation and translation of $\ba{T}$. 	Given the text parameters $\bv{\theta}$ and the observed text point (with homogenous coordinates $\tilde{\bv{m}}$ ) in the host image, the 3D coordinates of point $\bv{p}$ in the host frame are :	\beq
	\bv{p} = \tilde{\bv{m}}/\rho=\tilde{\bv{m}}/(\bv{\theta}^\vts \tilde{\bv{m}}).
	\eeq
	The point is then transformed into the target frame by 
	\beq
	\bv{p}' = \ba{R}\bv{p}+\bv{t}.
	\eeq
	Let $\tilde{\bv{m}}'$ be the homogeneous coordinates of the projected point on the target image plane. We have 
	\beq
	\begin{array}{c}
		\tilde{\bv{m}}' \sim \rho \, \bv{p}' = \ba{R} \tilde{\bv{m}} + \rho\bv{t} \\
		
		\Rightarrow \tilde{\bv{m}}' \sim \ba{R} \tilde{\bv{m}} + \bv{t}\, \bv{\theta}^\vts\tilde{\bv{m}},
	\end{array}
	\eeq
	where $\sim$ means equivalence up to a scale. Therefore, the process of projecting a 3D text object on the target image plane is a homography mapping of the text points from the host image plane to the target image plane, namely
	\beq
	\tilde{\bv{m}}' \sim \ba{H}\tilde{\bv{m}},
	\label{eq:homography}
	\eeq
	where $\ba{H} = \ba{R}+\bv{t}\bv{\theta}^\vts \in \mathbb{R}^{3\times3}$ is a homography matrix that relies on the relative pose $\ba{R},\bv{t}$ and the text parameters $\bv{\theta}$. 
	For convenience, we write the projection process as a function:
	\beq
	\bv{m}' = \bv{h}(\bv{m},\ba{T}_{h},\ba{T}_t,\bv{\theta}),
	\label{eq:text_prediction}
	\eeq
	where $\bv{m}$ denotes the observed text point on the host image plane, and $\bv{m}'$ represents the projected text point on the target image plane.  $\bv{\theta}$ represents the text parameters. 

	\anno{We take each text region as a single patch and align it to other frames directly by minimizing the difference between them instead of detecting the word once again.} 
	Motivated by directed approaches \cite{forster2014svo}\cite{engel2018direct}, our observation model computes the photometric error between the extracted text object and the projected one on the image. As we shall see in the experiments \anno{(see \Fig{fig:rapidData})}, using direct approaches will lead to better accuracy and robustness, particularly for blurry images. 
	The biggest issue of the direct approach is handling the illumination changes. Existing work \cite{engel2018direct} adopts an affine model to address intensity changes, but it requires extra parameters involved in optimization and sophisticated photometric calibration to guarantee performance.	
	We choose to use zero-mean normalized cross-correlation (ZNCC) as the matching cost to handle illumination changes. 
	
	Let $\Omega$ be the set of pixels within the text region, and $\bv{m} \in \Omega$ be a text pixel. The normalized intensities for text pixels are:
	\beq
	\tilde{{I}}(\bv{m}) = ({I}(\bv{m}) - \bar{{I}}_\Omega)/\sigma_\Omega,
	\label{eq:norm_image}
	\eeq
     where $\bar{{I}}_{\Omega}$ and $\sigma_{\Omega}$ stand for the average intensity and the standard deviation of the pixels in the text region $\Omega$.
	The text patch in the host image and the predicted one in the target image \Eq{eq:text_prediction} are then compared by :
	\beq
	ZNCC({{I}}_h,{{I}}_t) = \sum_{\bv{m}\in\Omega} \tilde{{I}}_h(\bv{m}) \tilde{{I}}_t(\bv{m}').
	\label{eq:zncc}
	\eeq
	The ZNCC cost is between $-1$ and $1$. The larger ZNCC cost indicates the two patches are more similar. However, it is difficult to directly use 
	the ZNCC cost within the optimization framework of the nonlinear least-squares problem. We hence adopt a variant form of ZNCC as the cost function
	\beq
	E({{I}}_h,{{I}}_t) = \sum_{\bv{m}\in\Omega} (\tilde{{I}}_h(\bv{m})-\tilde{{I}}_t(\bv{m}'))^2.
	\label{eq:norm_ssd}
	\eeq
	Though the cost function is similar to the SSD (Sum of Squared Difference) cost, it contains an additional normalization process to ensure the robustness to illumination changes. If we expand this cost function as : 
	\beq
	\sum_{\bv{m}\in\Omega}(\tilde{{I}}_h(\bv{m})^2 +\tilde{{I}}_t(\bv{m}')^2) - 2\sum_{\bv{m}\in\Omega}\tilde{{I}}_h(\bv{m}) \tilde{{I}}_t(\bv{m}'),
	\eeq
	we can discover that minimizing this cost function is equivalent to maximizing the ZNCC cost, because  $\sum\tilde{{I}}_h(\bv{m})^2 = N $ and $\sum\tilde{{I}}_t(\bv{m}')^2 = N$, where $N$ is a constant number of pixels within the text region $\Omega$. The photometric error of a text object $\pi$ with respect to the target frame $t$ is defined as :
	\beq
	E^{\pi,t}_{photo} = \sum_{\bv{m}\in\Omega^\pi} \phi((\tilde{{I}}_h(\bv{m})-\tilde{{I}}_t(\bv{h}(\bv{m},\ba{T}_h, \ba{T}_t, \bv{\theta}^\pi)))^2),
	\label{eq:photometric_error}
	\eeq
	where $\phi(\cdot)$ is the Huber loss function to handle possible outliers. Here, we use $\Omega^\pi$ to represent the text region on the host image plane. As we'll describe later, to make the computation faster, we do not use all the pixels within the text region, instead select only some of them as the reference pixels to compute the photometric error. 

\subsection{Semantic information management}
\label{sec:semantic_info_man}
	
	\begin{figure}[!tbp]
	\anno{
		\centering  
		\includegraphics[width=0.48\textwidth]{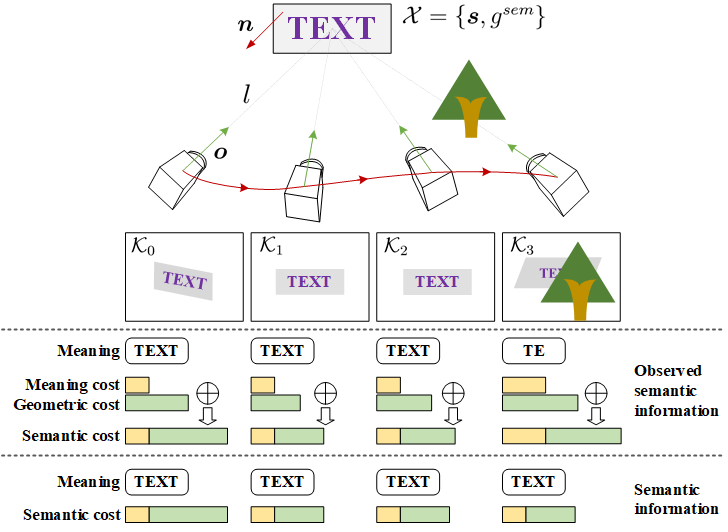}\\  
		\caption{The semantic information of a text object is updated whenever a new observation comes. \textbf{The top row} shows the text object and the camera trajectory as well as four keyframes. Note that in the fourth keyframe $\mathcal{K}_3$, the text object is occluded by the tree and is therefore partially observed. 
			\textbf{The second row} shows the observed semantic information extracted from each frame, which consists of the meaning of the text object (represented by a string) and the \anno{semantic costs} (including the meaning and geometric parts, the smaller the better). 
			\textbf{The third row} demonstrates the semantic information of the text object is updated from the observed one at each frame. The semantic information with \anno{the smallest semantic cost} is kept when updating. }  
		\label{fig:TextSLAMSemantic} 	
	}
	\end{figure}

	The semantic meanings of scene texts are valuable information for scene understanding and also benefit data association in SLAM because they are invariant to appearance changes.
	We represent the semantic information $\mathcal{X}$ of a text object by two parts: 
	its meaning $\bv{s}$ and \anno{a semantic cost} $g^{sem}$	
	\beq
	\mathcal{X} = \left\{\bv{s}, g^{sem}\right\},
	\eeq
	where the text meaning $\bv{s}$ is a text string and \anno{the semantic cost} $g^{sem}$ describes the quality of the estimated text meaning. \anno{Lower semantic costs} indicate better qualities.
	As illustrated in \Fig{fig:TextSLAMSemantic}, the semantic information of a text object is initialized from the first observation and continuously updated when new observations arrive. 
	
	For \anno{the text object recognized at each frame}, we extract its current semantic information, $\hat{\mathcal{X}} = \{\hat{\bv{s}},\hat{g}^{sem}\}$. Here $\hat{\bv{s}}$ is the text strings from the recognition results of the text extractor. The \anno{semantic cost} $\hat{g}^{sem}$ is defined as 
\beq
\hat{g}^{sem} = \lambda g^{mean} + g^{geo}.
\label{eq:semantic_cost}
\eeq
Here $g^{mean}$ represents \anno{the meaning cost with respect to} the confidence of the text extractor and $g^{geo}$ is \anno{a cost} describing if the text object is well posed towards the camera. The weight $\lambda$ is used to balance two sources of information, which is set as $200$ in our implementation. The smaller $\hat{g}^{sem}$ implies more reliable observed semantic information. 
	
\anno{The meaning cost $g^{mean}$ in (\ref{eq:semantic_cost}) is set as $g^{mean} = 1-g^{recg}$, where $g^{recg}$ comes from the confidence of text extraction\cite{zhang2019feasible}} and is usually located in the range of $[0,1]$. The larger confidence implies a more reliable recognition result. Here we use the minus operation to keep its consistency with other components. Some cases such as image blur or occlusion (take the fourth frame $\mathcal{K}_3$ in  \Fig{fig:TextSLAMSemantic} for example) will lead to a larger $g^{mean}$, indicating the extracted text meaning is unreliable.

\anno{The geometric cost $g^{geo}$ in (\ref{eq:semantic_cost})} is defined as 
$
g^{geo} =  l + \lambda'(1+\bv{o}^T\bv{n}/(\|\bv{o}\|\|\bv{n}\|)),
$
which consists of two terms. 
The first term measures the distance between the text object center and the camera center. The second term measures the difference between the viewing direction $\bv{o}$ and the normal direction $\bv{n}$ of the text object as visualized in \Fig{fig:TextSLAMSemantic}. The weight $\lambda'$  is set to $10$ in our implementation.

The semantic information of a newly detected text object is initialized as the first observation information $\mathcal{X}_0 \leftarrow \hat{\mathcal{X}}$. Whenever a new observation $\hat{\mathcal{X}}=\left\{\hat{\bv{s}}, \hat{g}^{sem}\right\}$ arrives, the semantic information of the text object 
	is updated by 
		\beq
	\mathcal{X}_k \leftarrow{
		\mathop{\arg\min}\limits_{\mathcal{X}_{k-1},\hat{\mathcal{X}}}
		(g^{sem}_{k-1}, \hat{g}^{sem}). 
	}
	\eeq
	In other words, \anno{the semantic information with the smallest semantic cost is selected}. 
	Based on this strategy, the extracted information under good conditions (legible and non-occluded text patches in the right orientation and close to the viewpoint) is preferred.
	Hence the semantic information will be more accurate with more \anno{high-quality} observations available.
	
\section{TextSLAM system}

\begin{figure}[!tbp]		
		\centering  
		\includegraphics[width=0.48\textwidth]{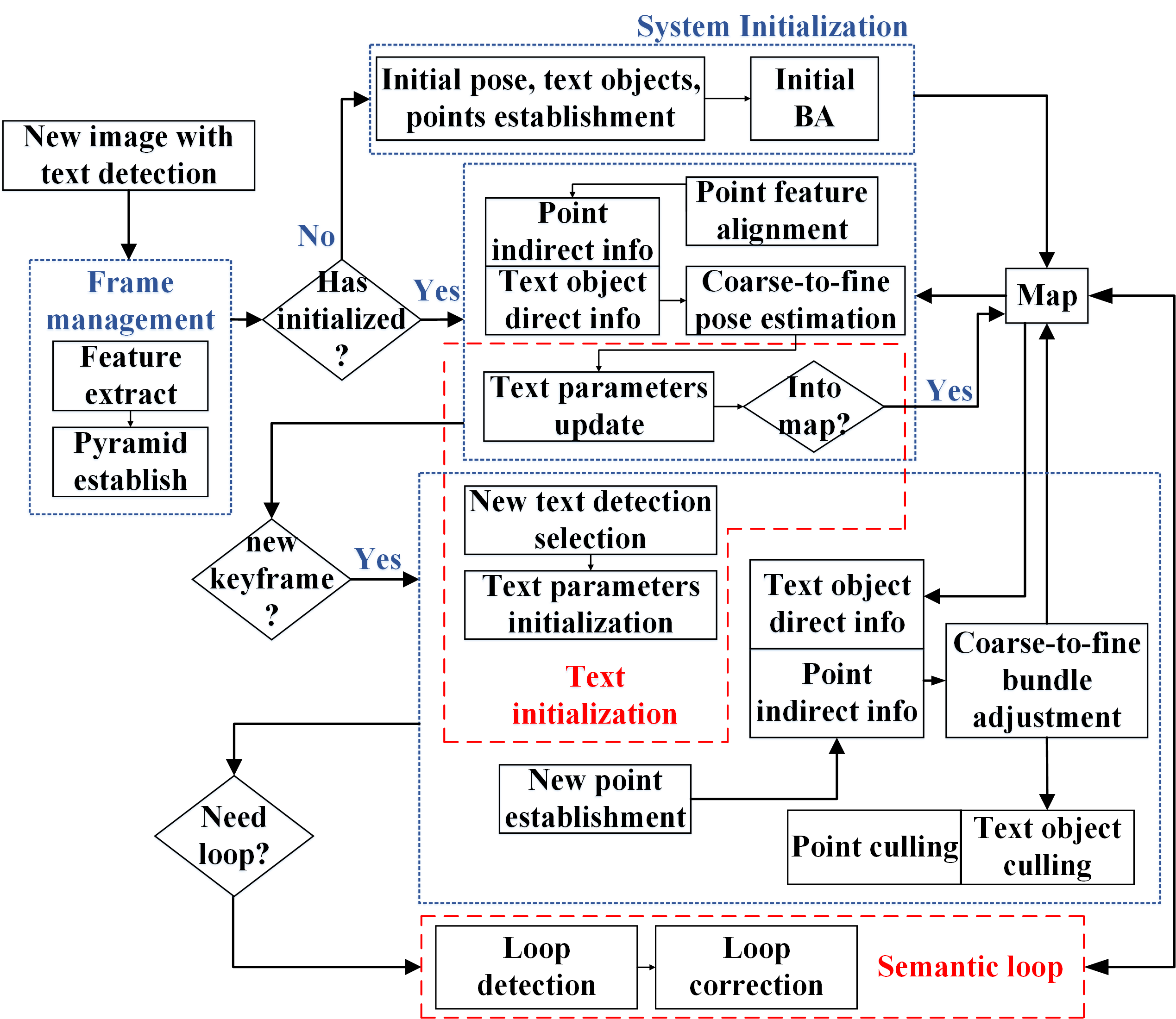}\\	
		\caption{An overview of our TextSLAM system.}  
		\label{fig:flowchart}  
	\end{figure}

	Our SLAM system, TextSLAM, is built upon a point-based system and adopts the keyframe-based framework to integrate the text features tightly. 
	The mixture of point features and text features allows our system to work properly even in the scenes without any text signs. 
	Fig. \ref{fig:flowchart} illustrates the flowchart of TextSLAM. We'll introduce the key components in the following sections.
	
\subsection{Initialization of text objects}
	\anno{A text object is initialized only when it is correctly extracted from the image.} Extracting scene texts from images is highly challenging until deep neural networks have been used. Recently, text detectors based on convolution neural networks \cite{zhang2019feasible, zhou2017east, lyu2018mask, wang2020all, liao2020real, he2021most, zhu2016scene, yin2016text} have achieved impressive performances.
	We adopt AttentionOCR\cite{zhang2019feasible} as the text extractor in our implementation, which ranks at the top on ICDAR2019 Robust Reading Challenge on Arbitrary-shaped Text \cite{chng2019icdar2019} and supports multiple languages. Some text extraction results are shown in \Fig{fig:BiJiang_Day_Map}. The outputs are arbitrary-orientation quadrilaterals enclosing the text regions. Note that our system is not limited to any particular text extractors. The text extractor can be replaced if more advanced ones are available.
	
	To initialize the parameters $\theta$ of a text object newly detected in a keyframe, we track the FAST \cite{rublee2011orb} feature points within the text region by Kanade-Lucas-Tomasi \cite{shi1993good} tracker until the next keyframe. Let $\bv{m}_i  \leftrightarrow \bv{m}'_i$ be the corresponding points in both keyframes, and $\ba{R},\bv{t}$ be the relative pose between the two frames. From \Eq{eq:homography}, we have $\tilde{\bv{m}}'_i\times \ba{H}\tilde{\bv{m}}_i = 0$. By taking $\ba{H} =  \ba{R}+\bv{t}\bv{\theta}^\vts$, we then obtain
	\beq
	[\tilde{\bv{m}}'_i]_\times \bv{t}\tilde{\bv{m}}^\vts_i \bv{\theta} =  -[\tilde{\bv{m}}'_i]_\times\ba{R}\tilde{\bv{m}}_i,
	\eeq
	where $\tilde{\bv{m}}_i$ and $\tilde{\bv{m}}'_i$ are the homogeneous coordinates of $\bv{m}_i$ and $\bv{m}'_i$ and $[\cdot]_\times$ denotes the skew symmetric matrix corresponding to the cross product. Note that the rank of the matrix on the left hand side is one. It requires at least three pairs of corresponding text points to solve $\bv{\theta}$.
	
After initialization of the parameters of a text object, we also keep the four corners of the quadrilateral indicating the text region.
The newly initialized text objects are kept being updated in the following frames. They are inserted into the 3D text map whenever the two rules are met: 1) the text object has been observed in at least $n_{min}$ ($4$ in our implementation) frames; 2) the text parameters converge to a relatively stable state. For the second rule, we check the normal of the text plane changes if larger than $25^\circ$ in our implementation. Once the parameters of a text object have been initialized, we also keep its semantic information being updated as described in \Sec{sec:semantic_info_man}. \anno{Several text map examples are visualized in \Fig{fig:textEx} and \Fig{fig:BiJiang_Day_Map}.}

	\anno{After successful initialization, each text is aligned to other frames directly by minimizing the photometric error.	
		Our method only requires sparse text detection to initialize texts that newly occurred and can track them reliably in the following frames without any extra detection inputs.} 
	
\subsection{Camera pose estimation with text objects}
	 
Both points and text objects are involved in camera pose estimation.
We select text objects observed by previous $2$ keyframes for pose estimation and exclude those that are behind the camera (at least one vertice of the text quadrilateral is behind the camera) or whose orientation is perpendicular to the current viewing direction. We also exclude those text objects whose appearance in the current frame changes much compared with that in the host frame \anno{because of occlusion}. We use ZNCC for comparison and exclude those text objects with ZNCC less than $0.1$ in our implementation. Camera pose estimation is done by minimizing the following cost function
\beq
E(\ba{T}_t) = E_{point}(\ba{T}_t) + \lambda_w E_{text}(\ba{T}_t),
\label{eq:pose_estimation_cost}
\eeq
where $\ba{T}_t \in SE(3)$ represents the current camera pose at frame $t$. The first term $E_{point}$ represents the sum of reprojection errors of point features : 
\beq
E_{point}(\ba{T}_t) = \sum_i \phi(\|\bv{m}_i-\mathcal{P}(\ba{T}_t, \bv{X}_i)\|^2),
\eeq
where $\bv{m}_i$ is the 2D coordinates of the observed point in the image and $\mathcal{P}(\ba{T}_t, \bv{X}_i)$ represents the projection of the 3D point $\bv{X}_i$ onto the image plane. Here $\phi(\cdot)$ is the Huber loss function to handle outliers. 
The second term $E_{text}$ contains only photometric errors of text objects, namely, 
\beq
E_{text} = \sum_{j} E^{\pi_j,t}_{photo}.
\eeq
Here $E^{\pi_j,t}_{photo}$ represents the photometric error of the $j$-th text object, which is defined in (\ref{eq:photometric_error}). Though we may use all the pixels within the text region to evaluate the photometric errors, an efficient way is to use a small part of them. Since the text region is full of textures, we adopt the FAST points\cite{rublee2011orb} within the text regions as the representative pixels. We then follow \cite{engel2017direct} to use an eight-pixel pattern around each representative pixel to compute the photometric errors.

The trade-off between the two terms in (\ref{eq:pose_estimation_cost}) needs to be regulated by the weight $\lambda_w$ since they are in different units (position difference vs intensity difference). The weight $\lambda_w$ is computed as $\lambda_w = \sigma_{rep}/\sigma_{photo}$. $\sigma_{rep}$ represents the standard deviation of the reprojection error of a pair of corresponding points (in both $x$ and $y$ directions) and $\sigma_{photo}$ represents the standard deviation of the photometric error of a text object as defined in \Eq{eq:norm_ssd}. Those standard deviations can be acquired through a small set of training data (given corresponding points and text patches).

Optimization of the cost function \Eq{eq:pose_estimation_cost} is a nonlinear least-squares problem. As the photometric cost $E_{text}$ is highly nonlinear, it requires a good initial guess of $\ba{T}_t$ to avoid being trapped in a local minimum. We firstly use a constant velocity model to predict the camera pose and then apply a coarse-to-fine strategy for efficient optimization.

Specifically, we downsize the images by $1/2$ recursively to build an image pyramid with three levels. Both the sampled points in the text region and detected feature points outside the text regions are down-sampled to reduce the number of variables to be optimized at coarse levels. The optimization starts from the coarsest level. The result is used to initialize the optimization process at the next level until reaching the final level. 
To downsample the text points in the next level, we divide the bounding box of a text object into a grid. For each cell in the grid, we select the point with the largest gradient. The number of cells for sampling is set to be $N_0 /4^l+100$, where $N_0$ is the number of points in the original resolution. Downsampling the feature points outside text regions works in a similar way, where the whole image is divided into cells for sampling.

During coarse-to-fine optimization, those points (including the text points) with large errors are marked as outliers and discarded. For each text object, when more than $99 \%$ text points are marked as outliers, this text object is marked as an outlier at this frame. 

\subsection{Text objects culling}	

To ensure the good quality of the 3D text map, we drop those text objects from further processing which have been frequently recognized as outliers in camera pose estimation.
Specifically, let $\#F_{bad}, \#F_{good}$ be the number of frames where the text object is marked and not marked as an outlier respectively. We check if the following conditions hold for the text object after finishing each bundle adjustment in our implementation:
\begin{enumerate}
	\item the text object is marked as an inlier in at least two frames ($\#F_{good} > 2$); 
	\item the number of bad frames is less than the number of good frames and also less than a preset limit ( $\#F_{bad} < 0.9\#F_{good}$ and $\#F_{bad} < 40$).
\end{enumerate}	
If one of those conditions is not met, the text object is set as 'bad object' and excluded from future processing.

\subsection{Bundle Adjustment with text objects}
We apply bundle adjustment from time to time in a local window of keyframes similar to existing SLAM systems \cite{mur2015orb}. The cost function of bundle adjustment also consists of the point part and the text part :
\beq
E(\bv{x}) = E_{point}(\bv{x}) + \lambda_w E_{text}(\bv{x}).
\label{eq:ba_cost}
\eeq
The cost function resembles that of camera pose estimation while involving more parameters to be optimized. The variable $\bv{x}$ includes the camera poses of keyframes in the local window, the inverse depth of point features, and the text parameters.
We also adopt a coarse-to-fine method to optimize \Eq{eq:ba_cost} as described in camera pose estimation.

%
%
%

\subsection{Loop closing using scene texts}
	
	
	Scene texts are reliable landmarks for place recognition because their meanings are invariant to changing illuminations or viewpoints. We present how to use scene texts to detect revisited places and also integrate them to correct the accumulated error as in our SLAM system. 

\subsubsection{Detection of loop candidates} 
	To detect possible loops, we need to compare the latest keyframe with old keyframes. Existing visual SLAM systems usually use the bag-of-visual-words vectors \cite{mur2017orb,galvez2012bags} for comparison. The visual words, clustered from the feature descriptors, rely on the image appearances that may change drastically, leading to false or missing loop detection. By contrast, the meanings extracted from those text objects - the text strings or the real words - will not change with image appearances. Hence our idea for loop closing is to use those 'real words' instead of 'visual words' for searching the similar keyframes. 
	
	Our searching process consists of two steps. 
	\anno{The first step is to match reliable words (have been refined by multiple covisible frames) observed in the latest keyframe to existing 3D text objects in the map. 
	Directly matching a 3D text map is far more efficient than matching the 2D detections on the historical frames because the latter requires much more comparisons due to the repeated observations of a single word, as discussed in Section 5.3.2.}
	The second step is to select loop candidates from the keyframes associated with those matched 3D text objects (note that the keyframes within the sliding window of bundle adjustment are excluded from selection). 
	
	To match a word in the query frame to a 3D text object in the map, we directly compare their meanings (text stings) $\bv{s}_i$, $\bv{s}_j$ by
	\beq
	s(\bv{s}_i, \bv{s}_j) = \frac{\max(|\bv{s}_i|, |\bv{s}_j|)-d(\bv{s}_i, \bv{s}_j)}{\max(|\bv{s}_i|, |\bv{s}_j|)} \in (0,1],
	\eeq
	where $|\bv{s}|$ is the length of a string $\bv{s}$ and $d(\bv{s}_i, \bv{s}_j)$ is the Levenshtein distance \cite{levenshtein1966binary} between two strings, \anno{which measures the minimum operations changing one string $\bv{s}_i$ to the other string $\bv{s}_j$, including deletion, insertion and substitution. For example, changing 'seed' to 'seek' needs 1 operation: substituting 'd' with 'k'. So the distance is 1.} 	
	The two strings are matched when the similarity score $s(\bv{s}_i, \bv{s}_j)$ is above a threshold. With such a similarity score,  it allows two strings to be matched even they are not exactly the same, which may happen when the text object is partially occluded or falsely recognized. The threshold of being matched or not is selected based on the best matching result. If one text object in the query frame is exactly matched to a text object in the map, $s(\bv{s}_i, \bv{s}_j)=1$, we require all the text objects to be exactly matched by setting the threshold to be $1$. Otherwise, we set the threshold proportional to the maximum matching score $s^{max}$ by $\max(\frac{2}{3} s^{max}, 0.35)$ to address partial occlusion or false recognition of text objects empirically, \anno{where $0.35$ served as the minimal threshold, dedicating the two texts are matched when at least one-third characters are same among the entirety.} This adaptive threshold scheme increases the robustness of our system \anno{in different scenes}.  	
	
	The candidate keyframes (the top ten are selected) for loop closing are selected from the keyframes associated with those matched text objects where the number of matched text objects is greater than a threshold $s_{min}$, which is set to be proportional to ($60\%$) the minimum number of covisibile text objects in the keyframe connected to the latest frame in the covisibility graph \cite{mur2015orb},  while being larger than three for outdoor scenes and two for indoor scenes in our experiments. 
	
	\begin{figure}[!h]
		\centering  
		\includegraphics[width=0.48\textwidth]{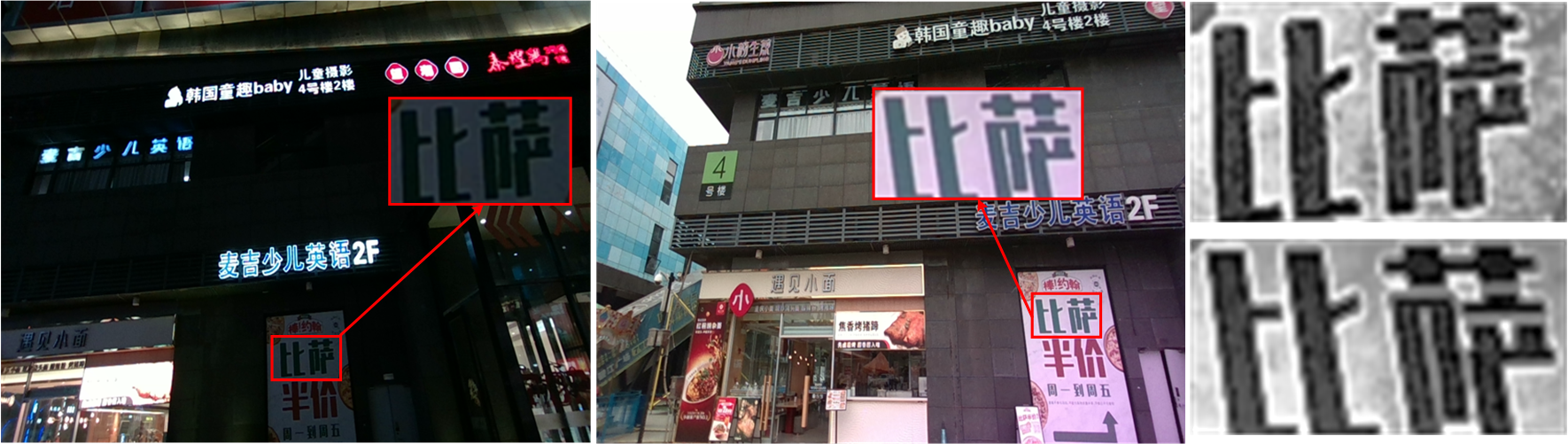}\\  
		\caption{\textbf{The first two columns} show the images captured at the same location in both day and night. The text sign with red rectangles is enlarged to show the dramatic appearance changes between day and night. \textbf{The third column} shows the text images after histogram equalization, where the top and the bottom are night and day images. Note that the contrast of the text patches does not change as much as we expect.}
		\label{fig:day_night_contrast}  
	\end{figure}

	\begin{figure}[!h]
		\centering  
		\includegraphics[width=0.48\textwidth]{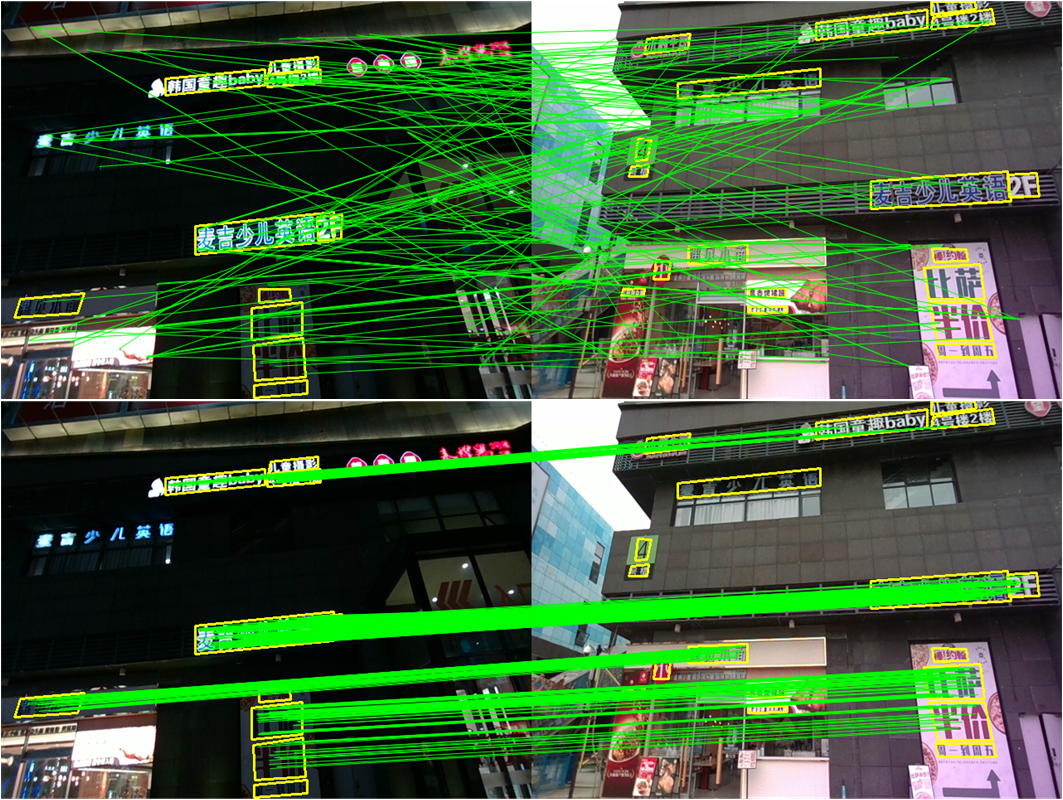}\\  
		\caption{\textbf{Top row}: The point correspondences by matching BRIEF descriptors.
			\textbf{Bottom row}: Result of our text-guided point matching.}
		\label{fig:SingleTextShow}  
	\end{figure}

	\begin{figure*}[!th]
		\centering  
		\includegraphics[width=1.0\textwidth]{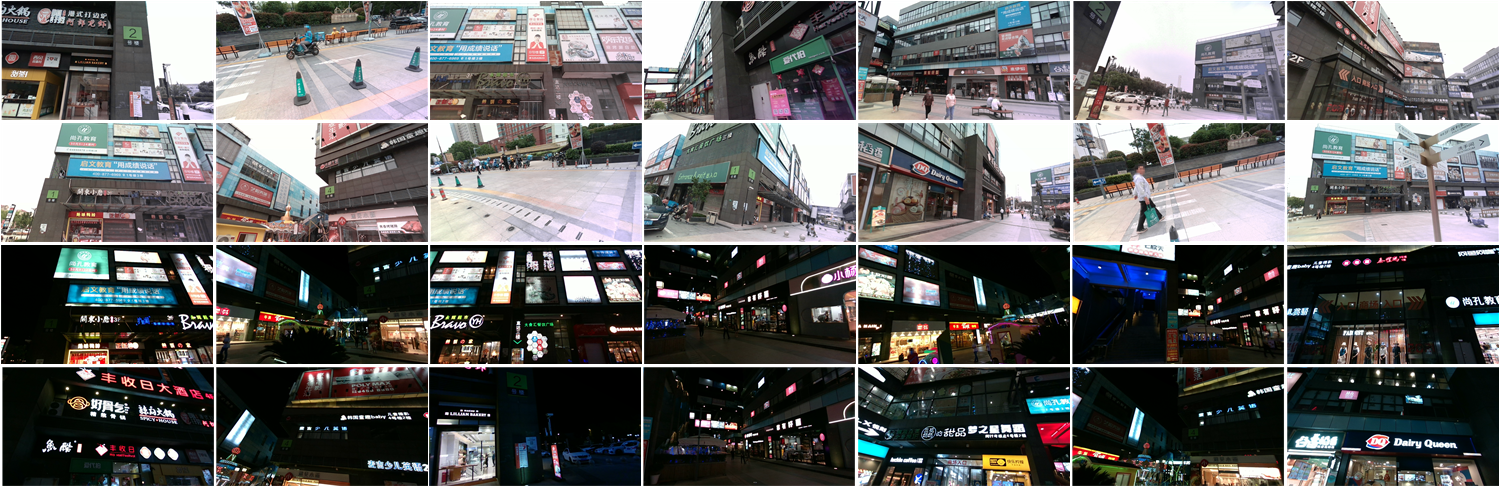}\\  
		\caption{Our outdoor datasets were collected in a commercial center, which is full of text signs with different sizes, fonts, and languages. The datasets consist of test sequences collected during both the day and night.}
		\label{fig:BiJiang_dataset}  
	\end{figure*}

\subsubsection{Compute the relative transformations} 
\begin{figure}[]		
	\centering  
	\includegraphics[width=0.48\textwidth]{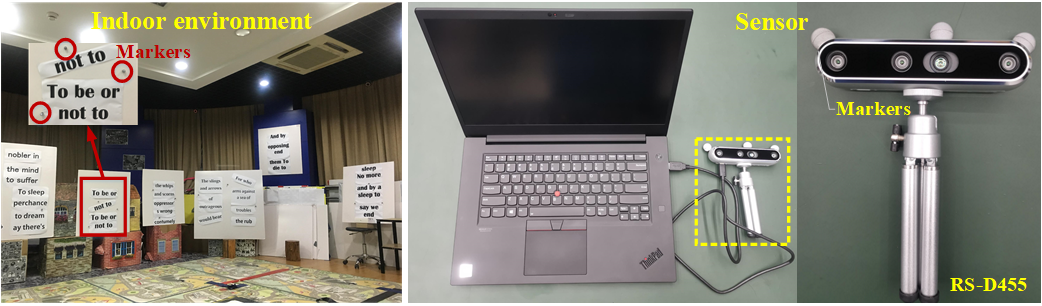}\\  
	\caption{The indoor test scene is shown on the left. The data collection device equipped with an RS-D455 camera is presented on the right.}  
	\label{fig:indoorScene_Sensor}  
\end{figure}

The relative transformation between the current keyframe and the loop frame is required to be estimated to close the loop. {We follow \cite{mur2015orb} to compute the similarity transformation between the current keyframe and the loop keyframe and the key is to obtain point correspondences between the two frames. However, it becomes highly challenging to acquire correct point correspondences when the illumination or viewpoint varies significantly. Since text objects are matched in loop detection by their semantic meanings, they can be used as a reliable prior for searching point-level correspondences even when illumination or viewpoint changes dramatically. Specifically, we search the point correspondences based on text points within the matched text regions instead of the whole images. We find that the contrast of a text object in the image does not change as much as we expect under different illuminations (unlike the color or intensity) as shown \Fig{fig:day_night_contrast}.
	Therefore, the BRIEF descriptor \cite{calonder2010brief}, relying on the relative difference of a pair of pixels, works well for matching the text points within the limited regions of two matched text signs, while it leads to a lot of false correspondences if matching is conducted on the whole images as shown in \Fig{fig:SingleTextShow}. 
	The text-guided point matching is robust and accurate even across the day and night as the experimental results show \anno{(\Fig{fig:BiJiang_Night_Traj} and \Fig{fig:localization_PointMatch_examples})}.

Similar to \cite{mur2015orb}, after we obtain the 3D to 3D correspondences from the matched text points, we use RANSAC to compute the similarity transformation and optimize it. Next, we perform a guided search to obtain more point correspondences outside the text regions. We then optimize the similarity transformation again and accept those loop candidates with sufficient inliers. 

\section{Experiments}	
	\begin{figure}[!h]
		\centering  
		\includegraphics[width=0.48\textwidth]{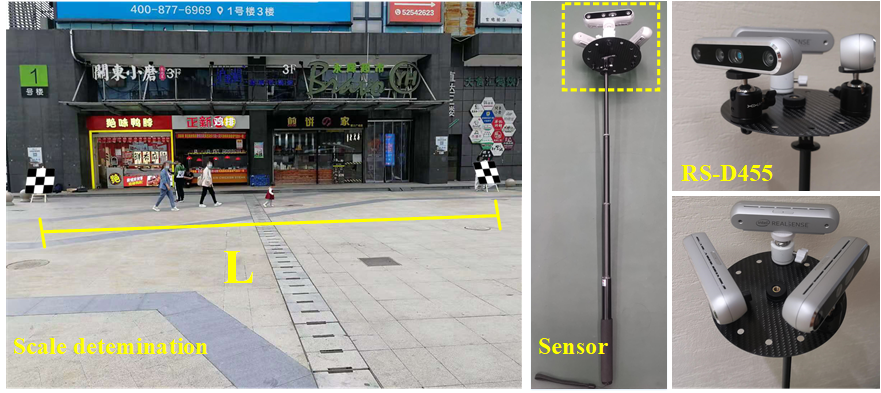}\\ 
		\caption{The outdoor test scene is shown on the left. The data collection device equipped with three RS-D455 cameras is presented on the right.}
		\label{fig:BiJiang_gt_scale_sensor}  
	\end{figure}

\subsection{Data collection}

	For the absence of SLAM benchmark datasets with text objects, we collected image sequences with scene texts in both indoor and outdoor scenes for evaluation.
	We use different devices for data collection in indoor and outdoor scenes.
	Our device for indoor scenes is shown in  \Fig{fig:indoorScene_Sensor}.
	It consists of \anno{an RGB camera} (Intel's RS-D455) for capturing the color images and several optical markers for obtaining ground truth via a motion capture system.
	Our device for outdoor scenes is shown in \Fig{fig:BiJiang_gt_scale_sensor}.
	It consists of \anno{three RGB cameras} recording multiple image sequences in different viewing directions simultaneously. We'll discuss how to acquire the ground truth trajectories for outdoors in the later sections. Image sequences are resized to $640 \times 480$ in all the tests. 
	
	
	
\subsection{Indoor tests}  
	
	Indoor tests were conducted within a laboratory. A room with a motion capture system was used to obtain the ground truth trajectories of millimeter accuracy as shown in \Fig{fig:indoorScene_Sensor}. The room was placed with random texts and those text strings were sampled from COCO-Text \cite{veit2016coco}, which is a large-scale text-orientated natural image dataset, where the fonts and sizes are randomly selected.

	We compare our TextSLAM system with the state-of-the-art visual SLAM systems: ORB-SLAM \cite{mur2017orb} and DSO\cite{engel2017direct}, where ORB-SLAM uses point features and DSO directly operates on raw pixels. By contrast, our system uses both points and text features and is not limited to text-rich scenes - if no text has been detected, our system can use only point features. 
	To evaluate the effectiveness of integrating text objects into the SLAM pipeline, we also present the results of our system with only point features enabled (Our point-only baseline).
	
\subsubsection{Evaluation of camera trajectory}

	\begin{table}[htbp]
		\caption{Results of indoor tests. RPE (1 m) and APE (m)}
		\begin{center}
			\scriptsize
			\setlength{\tabcolsep}{1.7mm}
			\begin{tabular}{ccccccccc}
				\hline	
				\multirow{2}{*}{\textbf{Seq.}} & \multicolumn{2}{c}{\textbf{ORB-SLAM}} & \multicolumn{2}{c}{\textbf{DSO}} & \multicolumn{2}{c}{\textbf{Our point-only}} & \multicolumn{2}{c}{\textbf{TextSLAM}} \\ \cline{2-9} 
				& \multicolumn{1}{c}{APE}         & \multicolumn{1}{c}{RPE}        & \multicolumn{1}{c}{APE}      & \multicolumn{1}{c}{RPE}      & \multicolumn{1}{c}{APE}          & \multicolumn{1}{c}{RPE}         & \multicolumn{1}{c}{APE}         & \multicolumn{1}{c}{RPE}        \\ \hline
				Indoor\_01   & 0.068 & 0.062  & 0.069 & 0.066  & 0.086 & 0.076  & \underline{{\bf 0.067}} & \underline{{\bf 0.055}} \\ 	
				Indoor\_02   & 0.092 & 0.075  & 0.070 & 0.060  & \underline{{\bf 0.067}} & 0.055   & 0.068 & \underline{{\bf 0.055}} \\	
				Indoor\_03   & 0.094 & 0.140  & 0.083 & 0.068  & 0.090 & \underline{{\bf 0.060}}  & \underline{{\bf 0.076}} &  0.111   \\	
				Indoor\_04   & 0.078 & 0.061  & 0.075 & 0.062  & \underline{{\bf 0.071}} & \underline{{\bf 0.047}}  & 0.076 & 0.060  \\	
				Indoor\_05   & 0.084 & 0.071  & 0.079 & 0.052  & 0.072 & 0.048  & \underline{{\bf 0.058}} & \underline{{\bf 0.037}}   \\ 
				Indoor\_06   & 0.089 & 0.070  & 0.074 & 0.054  & 0.084 & 0.059  & \underline{{\bf 0.073}} & \underline{{\bf 0.050 }}\\ 
				Indoor\_07   & 0.069 & 0.051  & 0.069 & 0.055  & 0.045 & \underline{{\bf 0.031 }}  & \underline{{\bf 0.032}} & 0.035\\ 
				Indoor\_08   & 0.081 & 0.077  & 0.068 & 0.055  & 0.051 & 0.046  & \underline{{\bf 0.049}} & \underline{{\bf 0.041}}  \\ 
				Indoor\_09   & 0.101 & 0.070  & {\bf 0.076} & {\bf 0.055}  & -- &  -- & \underline{0.096} & \underline{0.059}  \\	 
				Indoor\_10   & 0.075 & 0.062  & {\bf 0.071} & {\bf 0.047}  & 0.094 &  0.065 & \underline{0.076} & \underline{0.053 }\\ \hline 
			\end{tabular}
		\end{center}
		\emph{The middle bar '--' indicates the algorithm fails to finish the whole trajectory. The bold texts indicate the best results and the underlined texts highlight the better result between TextSLAM and the point-only baseline.}
		\label{tab:generalIndoor}
	\end{table}

In this test, we evaluate the performance of trajectory estimation of different systems by using the relative pose error (RPE) and the absolute pose error (APE).
Loop closing was disabled for all the systems. Ten indoor sequences were used for evaluation. The results are shown in \Tab{tab:generalIndoor}.
We can see that our TextSLAM system performs better than our point-only baseline, demonstrating the benefit of using the high-level text features.
	Our system also outperforms both ORBSLAM\cite{mur2017orb} and DSO\cite{engel2017direct} in most sequences and only performs slightly worse than DSO\cite{engel2017direct} in \anno{a few sequences (Indoor$\_09, 10$)}.
	Though the test scene with text labels is highly textured which is ideal for the point-based algorithms, our TextSLAM still performs the best among all the systems, again indicating the benefit from integrating text objects in the SLAM pipeline.
		
	We also evaluate the robustness of the proposed method under fast camera motion. Considering that commercial cameras, such as GoPro, are prone to image blur under fast motion, we use GoPro to collect image sequences under rapid motion. 
	The rapid motion causes severe image blur as shown in \Fig{fig:rapidData}, making our point-only baseline fail in all the cases. 
	ORB-SLAM works properly because of its well-implemented relocalization mechanism and it simply skips bad keyframes with image blur. By contrast, no relocalization is implemented in our system. Without the relocalization mechanism, DSO also fails in one test but still performs better than our point-only baseline. 
	By contrast, our text-based method works well in those tests. The text objects are successfully tracked as shown in \Fig{fig:rapidData} and the trajectories are more accurate than ORB-SLAM and DSO as shown in \Tab{tab:rapid}. This is largely due to tracking the text object as a whole by directly optimizing well-designed photometric errors.
	
	\begin{figure}		
		\centering  
		\includegraphics[width=0.49\textwidth]{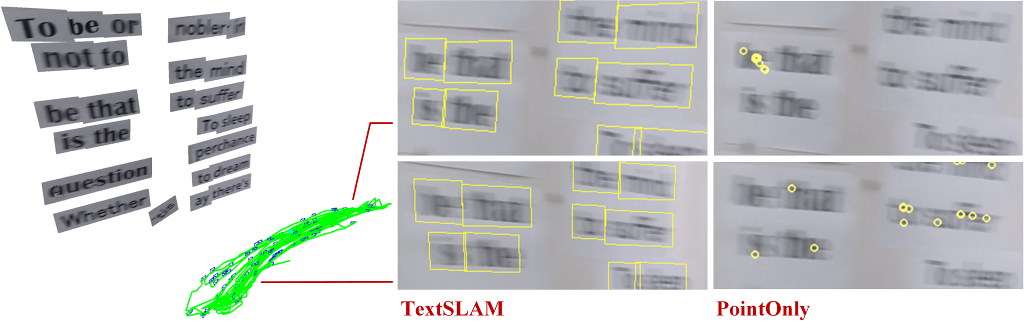}\\  
		\caption{TextSLAM is robust to blurry images caused by rapid camera motions. The estimated 3D text map and camera trajectory of TextSLAM are shown on the left. By contrast, the point-only method failed to track feature points on severely blurry images as shown on the right.}
		\label{fig:rapidData}  
	\end{figure}

	\begin{table}[]
		\caption{Results of rapid motion tests. RPE (0.1 m) and APE (m)}
		\begin{center}
			\scriptsize
			\setlength{\tabcolsep}{2mm}
			\begin{tabular}{ccccccccc}
				\hline
				\multirow{2}{*}{\textbf{Seq.}} & \multicolumn{2}{c}{\textbf{ORB-SLAM}} & \multicolumn{2}{c}{\textbf{DSO}} & \multicolumn{2}{c}{\textbf{Point-only}} & \multicolumn{2}{c}{\textbf{TextSLAM}} \\ \cline{2-9} 
				& \multicolumn{1}{c}{APE} & \multicolumn{1}{c}{RPE} &\multicolumn{1}{c}{APE} & \multicolumn{1}{c}{RPE} & \multicolumn{1}{c}{APE} & \multicolumn{1}{c}{RPE} & \multicolumn{1}{c}{APE} & \multicolumn{1}{c}{RPE} \\ \hline
				Rapid\_01 & 0.061 & 0.122 & -- & -- & -- & -- & {\bf 0.060} & {\bf 0.104} \\ 
				Rapid\_02 & 0.036 & 0.080 & 0.027 & {\bf 0.041} & -- & -- & {\bf 0.020} & 0.056 \\ 
				Rapid\_03 & 0.085 & 0.142 & 0.113 & {\bf 0.083} & -- & -- & {\bf 0.058} &0.107\\ \hline						
			\end{tabular}\\
		\end{center}
		\emph{The bar '--' indicates the algorithm fails to finish the whole trajectory.}
		\label{tab:rapid}
	\end{table}

\subsubsection{Evaluation of 3D Text Maps}
	Our TextSLAM system can directly produce a 3D text map. It would be interesting to evaluate the quality of the 3D text map. Since no other SLAM system generates text maps directly, we implement a baseline method by fitting the text planes to the 3D map points generated from ORB-SLAM and DSO within text regions \anno{using three-point RANSAC\cite{fischler1981random}}.
	
	To evaluate the quality of 3D text maps, we acquire the ground truth plane equation $(\bv{n}_{gt}, d_{gt})$ by placing optical markers on the text objects as shown in \Fig{fig:indoorScene_Sensor}. We use the angular error and the distance error between the estimated text plane and the ground truth for evaluation. The angular error measures the difference between the estimated normal $\bv{n}_{t}$ of the text plane and the ground truth $\bv{n}_{gt}$, namely $\alpha = \arccos(|\bv{n}_{t}^\vts\bv{n}_{gt}|/\|\bv{n}_{t}\|\|\bv{n}_{gt}\|)$. The distance error is measured by the distance between the 3D text points and the ground truth text planes, \anno{where the four corners of the text quadrilateral are selected as the 3D text points.}
	
	\anno{The statistics distributions of angular errors and distance errors over ten sequences are shown in \Fig{fig:3DTextMapping_statistic} and \Fig{fig:3DTextMapping_statistic_dist}, respectively. 
	The visual comparison of one test sequence is presented in \Fig{fig:3DTextMapping_visual}.}
	The results show that the 3D text map produced by TextSLAM is better than that of the plane fitting baselines based on either ORB-SLAM or DSO. This is largely due to the parameters of a text object being estimated as a whole in TextSLAM, while in ORB-SLAM or DSO, text points are estimated separately without considering they are on the same plane. Therefore, the fitted planes are noisy as shown in \Fig{fig:3DTextMapping_visual}.

	\begin{figure}[!h]		
		\centering  
		\includegraphics[width=0.45\textwidth]{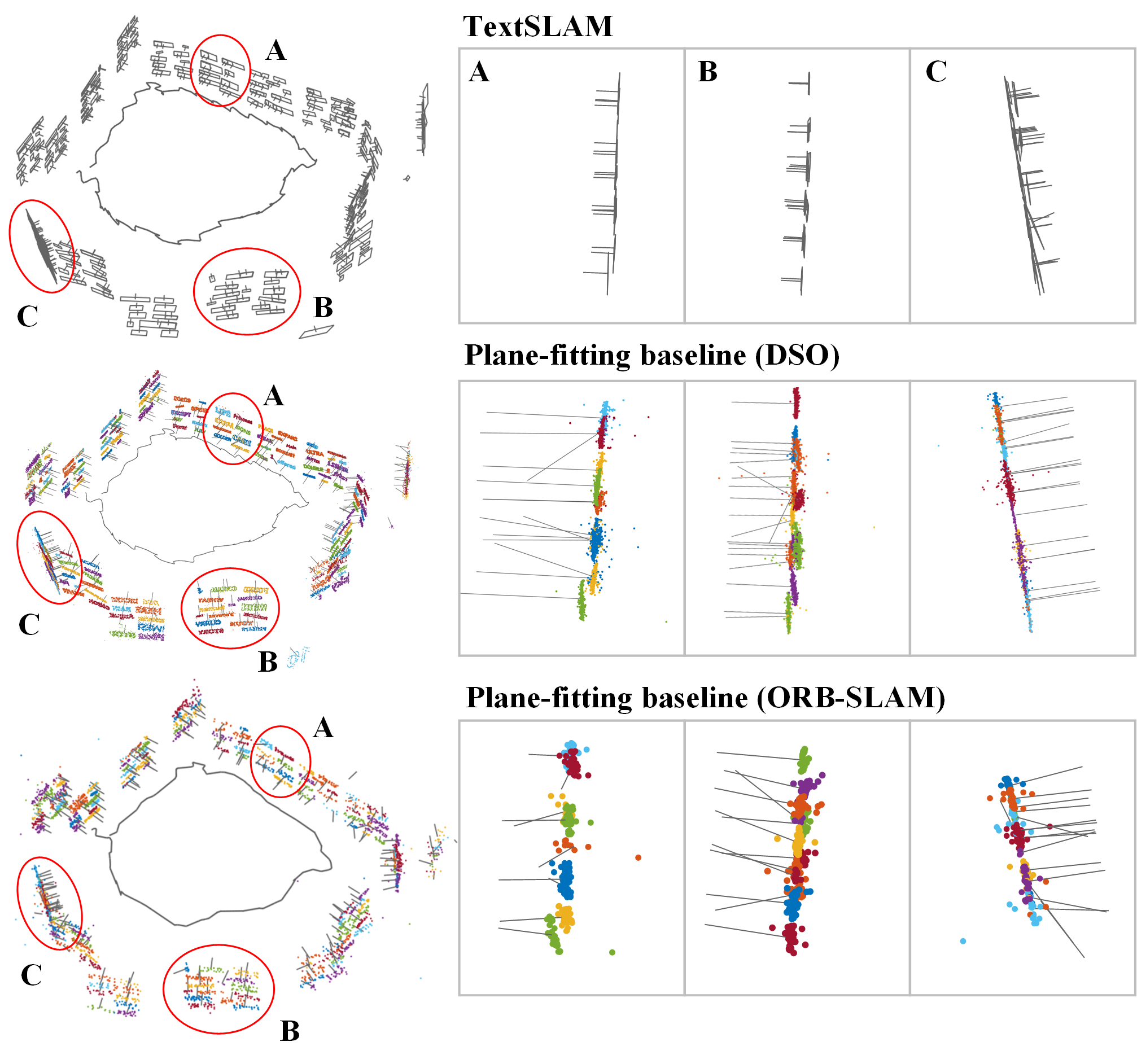}\\   
		\caption{
			Though RANSAC was adopted, plane fitting on the point clouds from ORB-SLAM still produced noisy results (as shown in the bottom row). DSO performs better because much more points were taken into the computation. By contrast, TextSLAM avoids such problems by tracking a text object as a whole via directly optimizing the photometric errors.}
		
		\label{fig:3DTextMapping_visual}  
	\end{figure}
	
	\begin{figure}[!h]
		\centering  
		\includegraphics[width=0.5\textwidth]{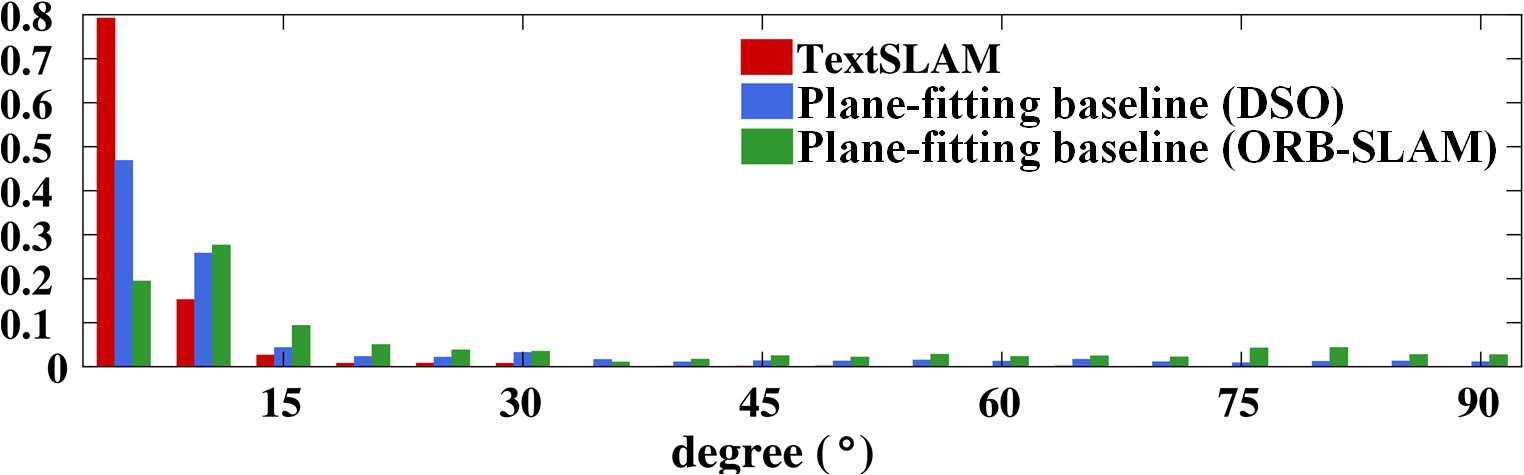}\\
		\caption{The statistic distribution of the angular errors.
			The results of TextSLAM, plane-fitting baselines based on DSO and ORB-SLAM are illustrated in red, blue, and green respectively.}
		\label{fig:3DTextMapping_statistic}  
	\end{figure}

	\begin{figure}[!h]
		\centering  
		\includegraphics[width=0.48\textwidth]{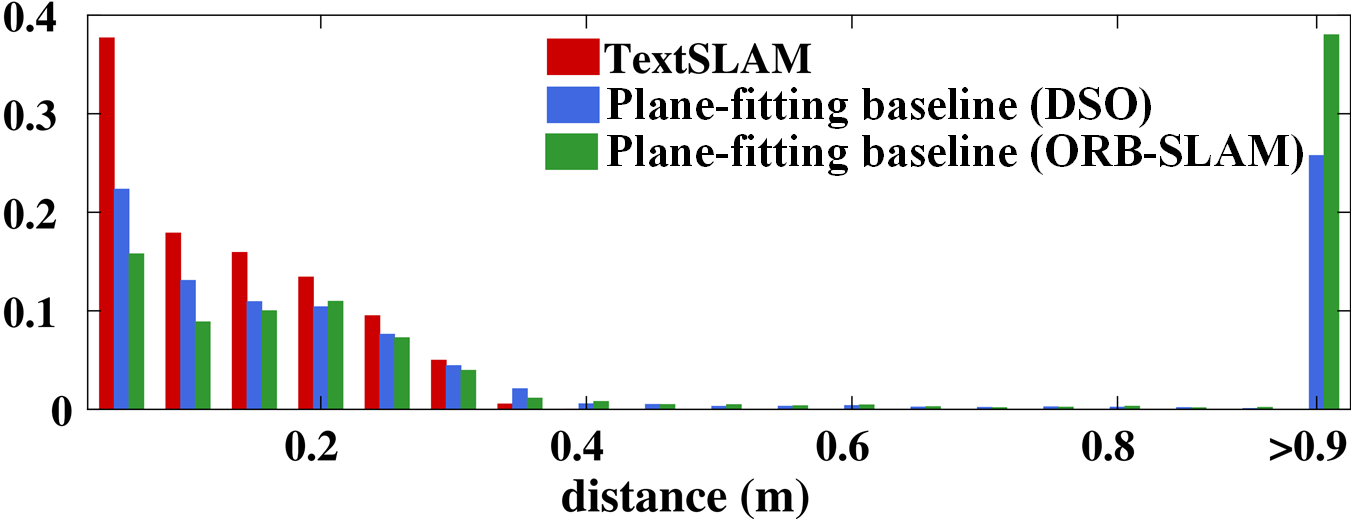}\\ 
		\caption{The statistic distribution of the distance errors. The results of TextSLAM, plane-fitting baselines based on DSO and ORB-SLAM are illustrated in red, blue, and green respectively.}
		\label{fig:3DTextMapping_statistic_dist}  
	\end{figure}

\subsubsection{Evaluation of loop closing}
	%
	\begin{figure}[!h]
		\centering  
		\includegraphics[width=0.48\textwidth]{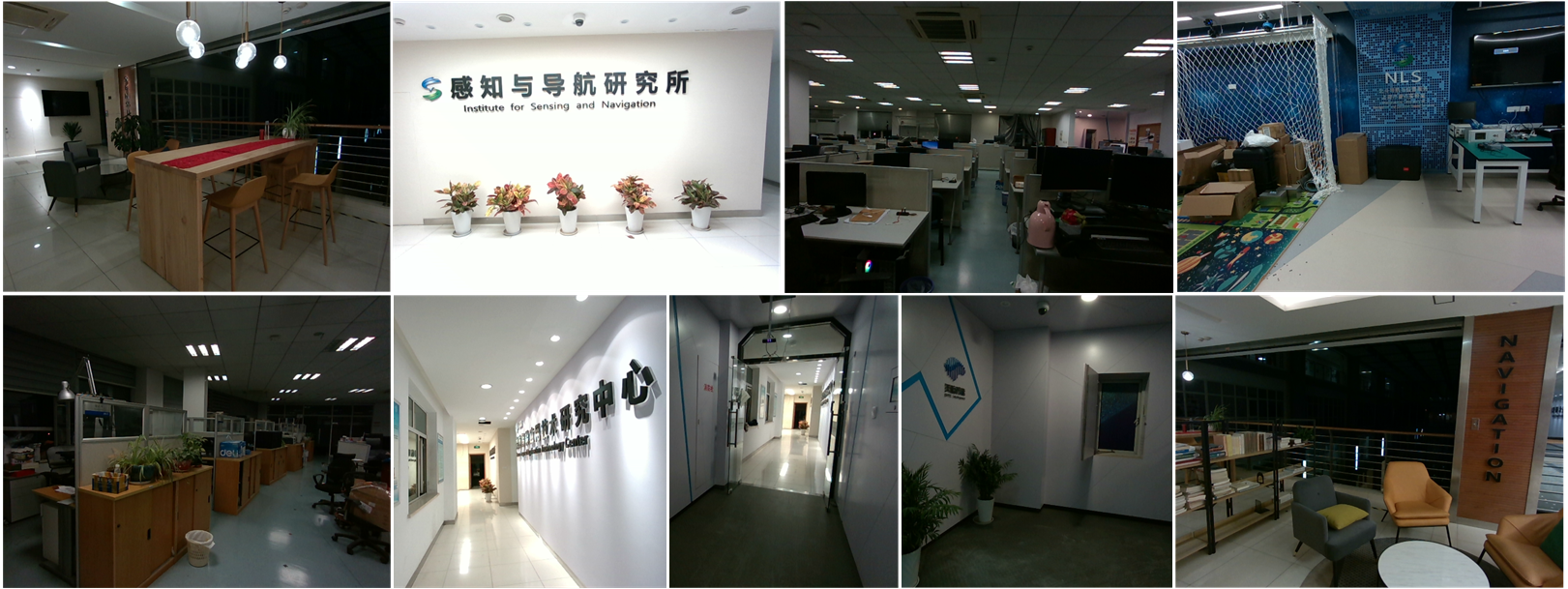}\\  
		\caption{The indoor datasets used for loop closing tests were collected within a laboratory environment where some sparse text signs are available. We collected the test sequences in day and night and also turned on and off the lights to make the tests more challenging.}
		\label{fig:indoorLoop2_env}  
	\end{figure}

	Additional sequences were recorded to evaluate the loop closing performance within two indoor scenes. The first scene is in a small room equipped with a motion capture system that provides the ground truth trajectories all the time. Some printed texts were randomly placed in the room similar to the first experiment as shown in \Fig{fig:indoorScene_Sensor}. The second scene spans the whole floor of a building and contains some sparse text signs or labels as shown in \Fig{fig:indoorLoop2_env}. We started and ended recording within the same room with the motion capture system to acquire the ground truth poses in the start and end parts of a trajectory. We follow \cite{zou2019structvio} to compute the positional errors using the partially available ground truth.
	
The results are shown in \Tab{tab:IndoorLoop1} and \Tab{tab:IndoorLoop2}}, where the ORB-SLAM's results are also presented for comparison. We also visualize some results in \Fig{fig:IndoorLoop1_visual} and \Fig{fig:IndoorLoop2_visual}.
	In those tests, ORB-SLAM failed to detect most loops, while TextSLAM can detect all the loops correctly. The reason is that the viewpoint of the current frame changes significantly from that of the loop frame (see \Fig{fig:IndoorLoop1_visual} and \Fig{fig:IndoorLoop2_visual}), making ORB features difficult to be matched. By contrast, our TextSLAM uses the semantic meaning of those text objects to detect loop frames. The semantic meaning of a text object will keep unchanged with viewpoint changes. The results suggest that text objects can be used as reliable landmarks for loop closing even though they are distributed sparsely in the scene. 
	
	\begin{table}[htbp]
		\caption{Loop tests in a small indoor scene. RPE (1 m) and APE (m)}
		\begin{center}
			\begin{tabular}{ccccccc}
				\hline
				\multirow{2}{*}{\textbf{Seq.}} & \multicolumn{3}{c}{\textbf{ORB-SLAM}} &  \multicolumn{3}{c}{\textbf{TextSLAM}} \\ \cline{2-7} 
				& LOOP		& APE         & RPE        & LOOP	& APE         & RPE        \\ \hline
				AIndoorLoop\_01 & $\surd$& {\bf 0.005} & 0.046 & $\surd$& 0.010 &  {\bf 0.031}\\	
				AIndoorLoop\_02 & $\times$& 0.075 & 0.077 & $\surd$& {\bf 0.018} & {\bf 0.030} \\	
				AIndoorLoop\_03 & $\times$& 0.076 & 0.090 & $\surd$& {\bf 0.007} & {\bf 0.042} \\	
				AIndoorLoop\_04 & $\surd$& 0.028 & 0.035 & $\surd$& {\bf 0.026} & {\bf 0.027}\\	
				AIndoorLoop\_05 & $\surd$& {\bf 0.008} & {\bf 0.034} & $\surd$& 0.017 & 0.040 \\	
				AIndoorLoop\_06 & $\times$& 0.068 & 0.069 & $\surd$& {\bf 0.010} & {\bf 0.027} \\	
				AIndoorLoop\_07 & $\times$& 0.083 & 0.071 & $\surd$& {\bf 0.017} & {\bf 0.032} \\	
				AIndoorLoop\_08 & $\times$& 0.082 & 0.086 & $\surd$& {\bf 0.008} & {\bf 0.026} \\ \hline	
			\end{tabular}
		\end{center}
		 \emph{The tick '$\surd$' indicates a success loop closing and '$\times$' indicates no loop has been found. The smallest errors are in bold texts.}
		\label{tab:IndoorLoop1}
	\end{table}

	\begin{figure}[!h]
		\centering  
		\includegraphics[width=0.48\textwidth]{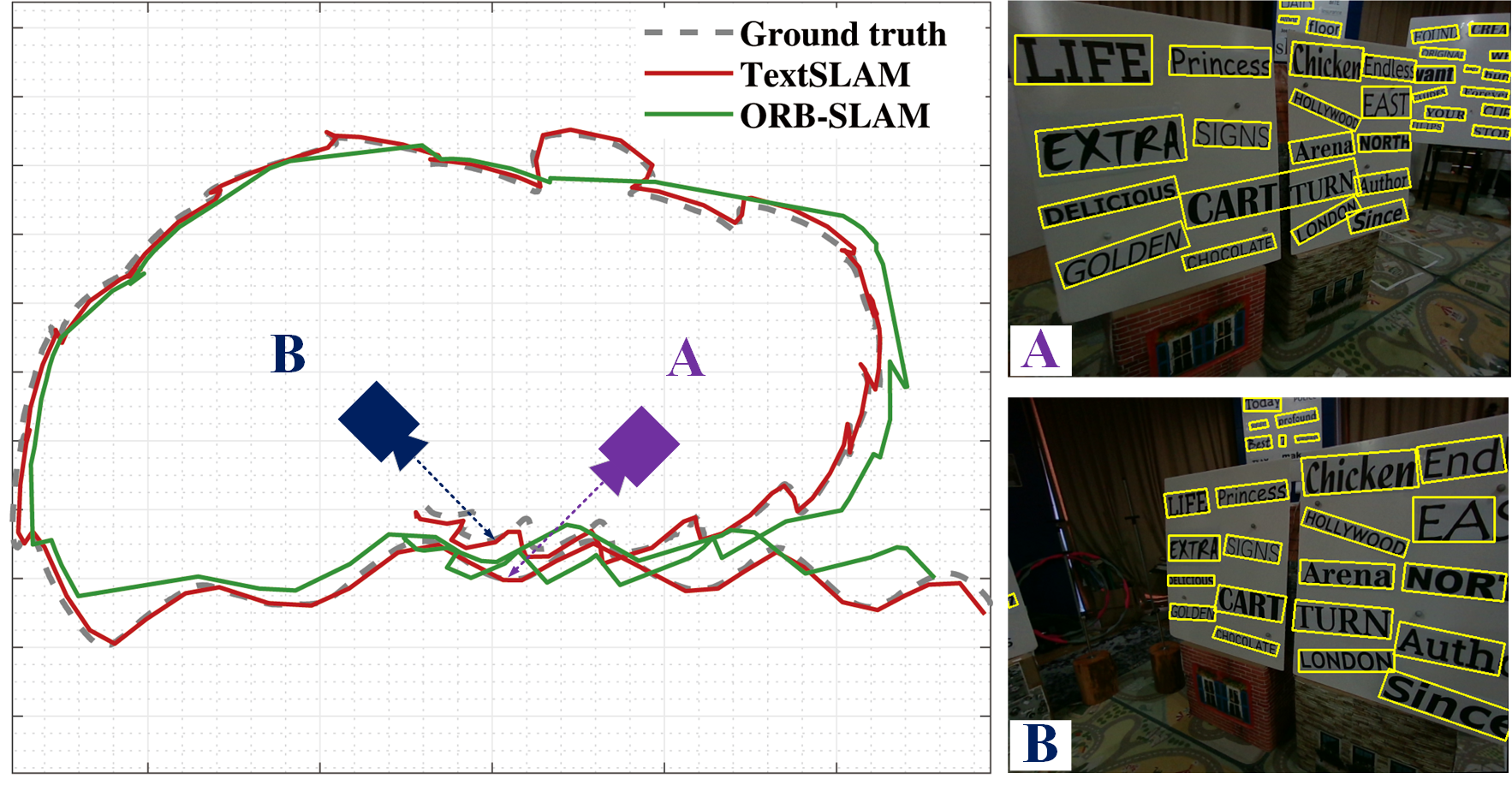}\\  
		\caption{Visualization of TextSLAM results in a small indoor scene. The trajectories of the two methods are shown on the left. The results of TextSLAM, ORB-SLAM, and the ground truth are illustrated in red, green, and gray respectively. The query frame $B$ and the detected loop frame $A$ are shown on the right, where the viewpoint changes significantly.}
		\label{fig:IndoorLoop1_visual}  
	\end{figure}

	\begin{table}[htbp]
		\caption{Loop test in a large indoor scene. RPE (1 m) and APE (m)}
		\begin{center}
			\begin{tabular}{ccccccc}
				\hline
				\multirow{2}{*}{\textbf{Seq.}} & \multicolumn{3}{c}{\textbf{ORB-SLAM}} &  \multicolumn{3}{c}{\textbf{TextSLAM}} \\ \cline{2-7} 
				& LOOP		& APE         & RPE        & LOOP	& APE         & RPE        \\ \hline
				LIndoorLoop\_01 & $\times$& 0.994 & 0.770 & $\surd$& {\bf 0.062} & {\bf 0.111} \\ 
				LIndoorLoop\_02 & $\times$& 1.669 & 1.177 & $\surd$& {\bf 0.057} & {\bf 0.071} \\ 
				LIndoorLoop\_03 & $\times$& 2.102 & 0.595 & $\surd$& {\bf 0.192} & {\bf 0.374} \\ 
				LIndoorLoop\_04 & $\times$& 0.192 & 0.202 & $\surd$& {\bf 0.023} & {\bf 0.075} \\ 
				LIndoorLoop\_05 & $\times$& 0.251 & 0.127 & $\surd$& {\bf 0.047} & {\bf 0.065} \\ 
				LIndoorLoop\_06 & $\times$& 0.179 & 0.163 & $\surd$& {\bf 0.032} & {\bf 0.041} \\ 
				LIndoorLoop\_07 & $\times$& 0.206 & 0.328 & $\surd$& {\bf 0.031} & {\bf 0.230} \\ 
				LIndoorLoop\_08 & $\times$& 0.291 & 0.155 & $\surd$& {\bf 0.031} & {\bf 0.042} \\ 
				LIndoorLoop\_09 & $\times$& 0.377 & 0.202 & $\surd$& {\bf 0.031} & {\bf 0.034} \\ 
				\hline
			\end{tabular}
		\end{center}
		\emph{The tick '$\surd$' indicates a success loop closing and '$\times$' indicates no loop has been found. The smallest errors are in bold texts.}
		\label{tab:IndoorLoop2}
	\end{table}
	
	\begin{figure*}[!h]
	\anno{
		\centering  
		\includegraphics[width=1\textwidth]{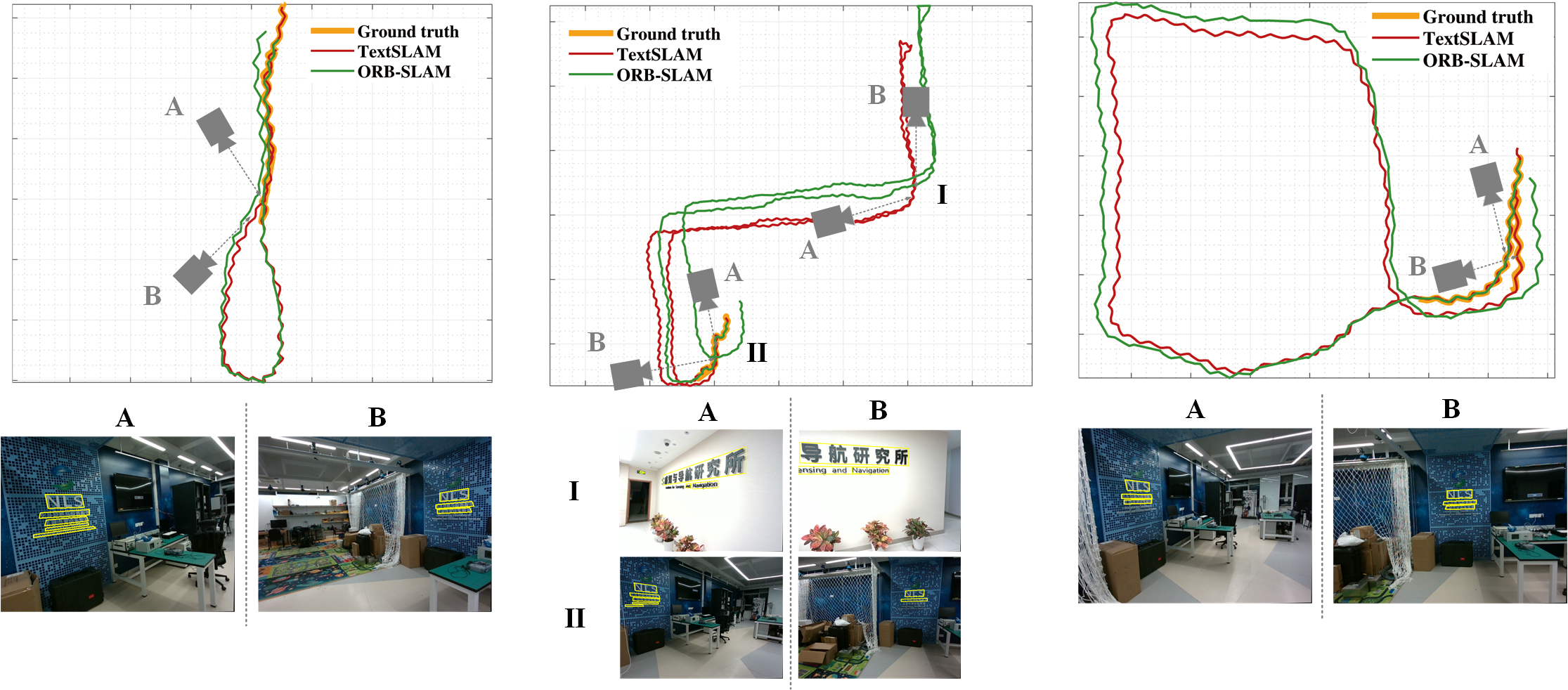}\\  
		\caption{Visualization of TextSLAM results in a large indoor scene. \textbf{Top row:} The camera trajectories of TextSLAM, ORB-SLAM, and the ground truth are visualized in red, green, and orange respectively. The ground truth (orange line) is at \textbf{the start and end parts} of each trajectory. \textbf{Bottom row:} The query frame $B$ and the detected loop frame $A$ in TextSLAM are visualized, where the matched text objects are highlighted in yellow boxes. We can see that loops are correctly detected in TextSLAM despite large viewpoint changes.}
		\label{fig:IndoorLoop2_visual}  
	}
	\end{figure*}

\subsection{Outdoor tests}	

	\begin{figure}[!hb]
		\centering  
		\includegraphics[width=0.5\textwidth]{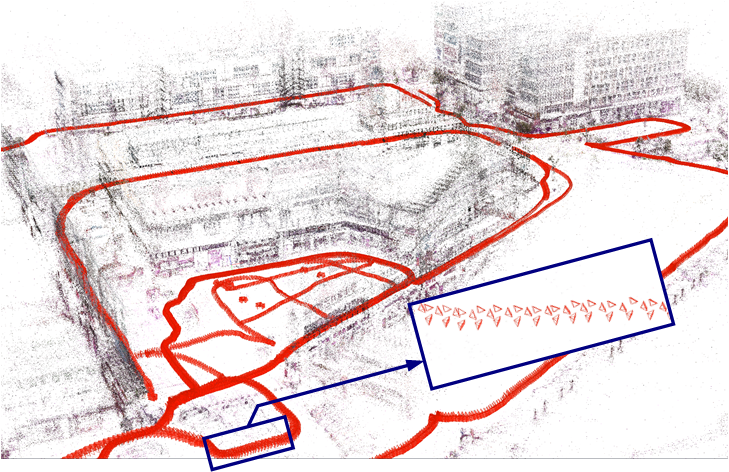}\\ 
		\caption{The structure-from-motion model that we used as ground truth in the outdoor tests. \anno{Three surround-view cameras} were used for data collection as illustrated in the enlarged area.}
			\label{fig:BiJiang_gt}  
	\end{figure}

	In this experiment, we test our TextSLAM system in a commercial plaza during the day and night. Some pictures are shown in \Fig{fig:BiJiang_dataset}. As we can see, 
	the environment is full of text objects with various sizes, fonts, backgrounds, and languages, as well as various challenges including complex occlusions, the reflection of the glass, and moving pedestrians.
	
	
	
	The ground truth camera trajectories are required for evaluation. One possible solution is to use the RTK GPS receiver to obtain the camera trajectories in centimeter-level accuracy. However, it is not feasible because we found that satellite signals were occluded by the surrounding buildings. Instead, we use the struct-from-motion technique to obtain the ground truth following the idea of \cite{Sattler_2017_CVPR, Sattler_2018_CVPR}. We collected a full set of image sequences to densely cover the scene and ran COLMAP \cite{schonberger2016structure} to obtain the camera pose for each image. After that, the camera poses obtained from COLMAP are treated as the ground truth and used to evaluate the SLAM performance. \anno{The 3D map and camera trajectories from COLMAP of the outdoor scene are visualized in \Fig{fig:BiJiang_gt}.} We selected eight sequences among the full set of image sequences for evaluation. To cover the scene more efficiently, we use three cameras with different headings to capture the images in different viewpoints as shown in \Fig{fig:BiJiang_gt_scale_sensor}. 
	
	Since COLMAP produces 3D structures with an unknown scale, we need to calibrate the scale by a reference distance. As shown in \Fig{fig:BiJiang_gt_scale_sensor}, we placed two checkerboards in the scene. Their orientation was kept the same such that the distances between corresponding points on the board are identical. The checkerboard corners can be extracted and matched, whose 3D coordinates can be estimated from the known camera poses produced by structure-from-motion. We compared the estimated distance from structure-from-motion and the measured distance by a laser rangefinder to resolve the unknown scale. 
	
	
	\anno{To evaluate the accuracy of the ground truth, we chose five reference points whose real-world coordinates were precisely measured by a laser rangefinder. We then placed the camera at those reference points to capture extra images and fed them into the COLMAP pipeline. Those estimated locations were aligned with the real-world coordinates to evaluate the accuracy of the ground truth approximately. We found that the average localization error is about $8.45 cm$ within the test area around $5500 m^2$, which is sufficient for our evaluation.} 

\subsubsection{Day tests}
\begin{figure*}[!h]
		\anno{
		\centering  
		\includegraphics[width=\textwidth]{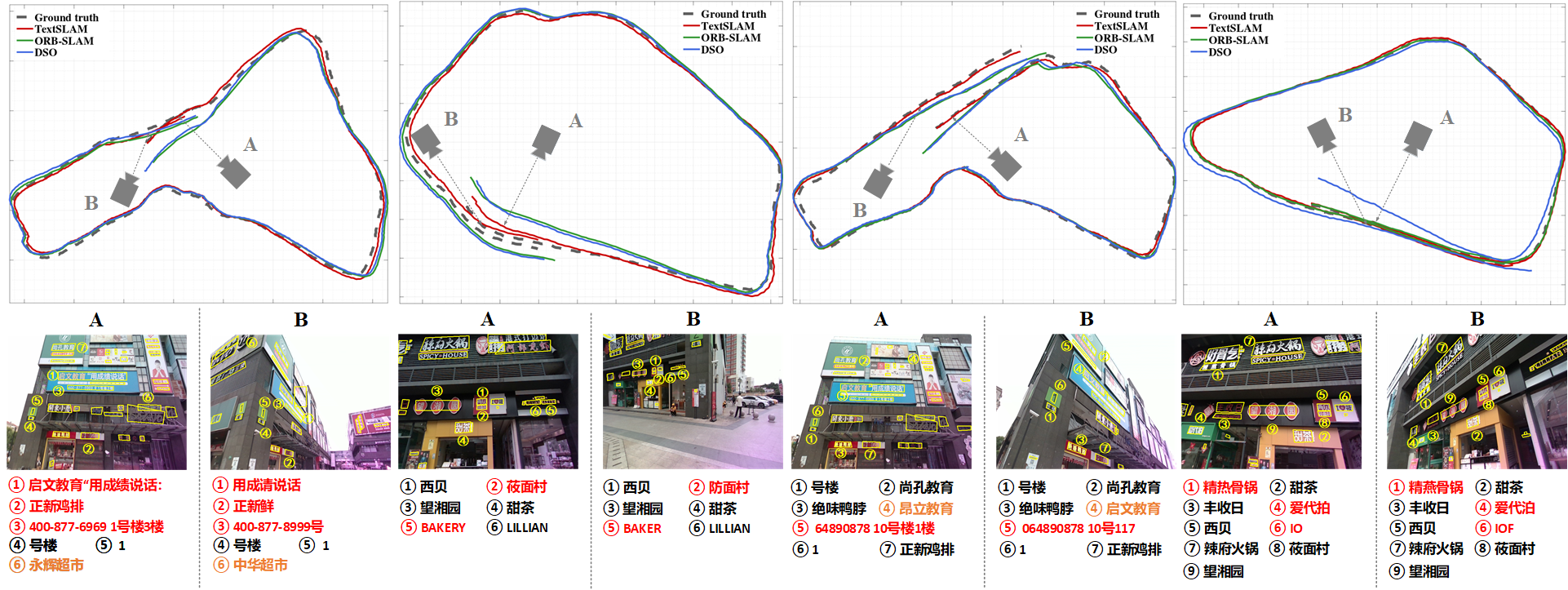}\\  
		\caption{Results of outdoor tests.\textbf{Top row:} The camera trajectories of TextSLAM, ORB-SLAM, DSO, and the ground truth are visualized in red, green, blue, and gray respectively. The loop frames detected by TextSLAM are also visualized ($B$ represents the query frame and $A$ is the detected loop frame).
			 \textbf{Bottom row:} The semantic meanings of those matched text objects between the loop frame and the query frame are presented, where the matched pair are indicated by the same number. 
			Our method allows those strings to be exactly matched (in black) or partially matched (in red). Some false matching results are shown in brown, which are excluded from the geometric verification during loop closing.}	
		\label{fig:BiJiang_Day_Traj}  
		}
	\end{figure*}

\begin{figure*}[!h]
		\centering  
		\includegraphics[width=\textwidth]{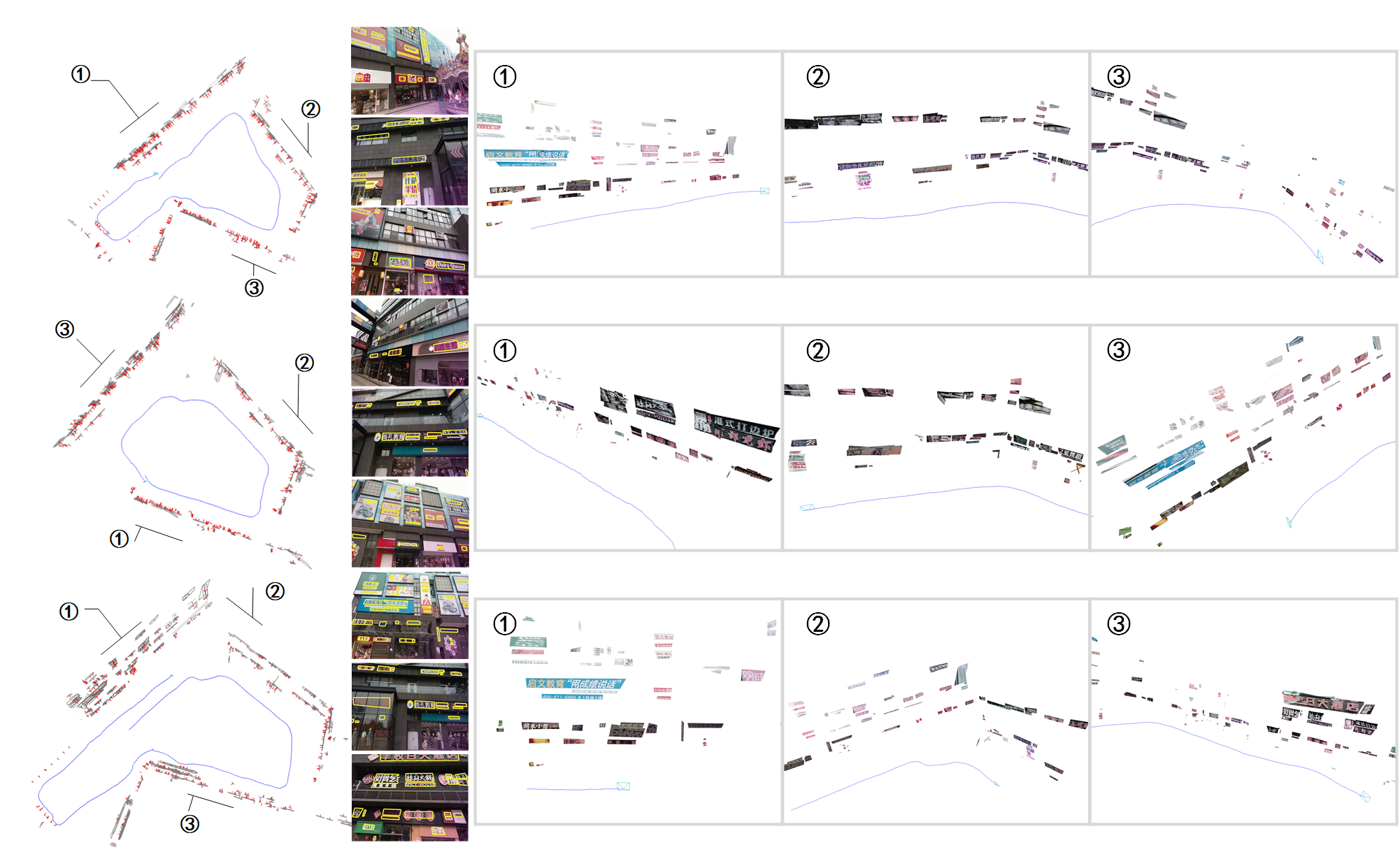}\\  
		\caption{Extra results of outdoor tests. \textbf{First column:} The full mapping and localization results are shown. \textbf{Second column:} The text detection results of three numbered locations are visualized. \textbf{Third to fifth columns:} \anno{The 3D text map} in marked locations are zoomed in for more details.}
		\label{fig:BiJiang_Day_Map}  
	\end{figure*}

	In this experiment, we evaluate our methods with the image sequences collected during the day. We also present the results of ORB-SLAM and DSO for comparison.
	The results are shown in \Tab{tab:BiJiang_DayTraj}.   
	TextSLAM can correctly recognize revisited places and close the loop in all test sequences, achieving the best accuracy among all the methods.
	ORB-SLAM fails to detect most loops because of large viewpoint changes and performs similar to DSO that has no loop closing function. 
	To be more clear, we visualize estimated trajectories for typical sequences in \Fig{fig:BiJiang_Day_Traj}, as well as the loop image pairs with text objects detected by TextSLAM. As shown in \Fig{fig:BiJiang_Day_Traj}, the viewpoints change significantly between the current keyframe and the loop keyframe, making the BoW-based method (ORB-SLAM) fail to detect those loops.
	By contrast, the semantic message of a text object is invariant to appearance changes, hence TextSLAM is able to detect the correct loop via using this high-level information.
	When the viewpoint changes only slightly, e.g. Outdoor\_5 in \Tab{tab:BiJiang_DayTraj}, the well-implemented ORB-SLAM can correctly close the loop and produce results as accurate as ours. 
	Extra results of TextSLAM are presented in \Fig{fig:BiJiang_Day_Map}, including the reconstructed 3D text map for all scene texts existing in the test scene, as well as their 2D observations. The 3D semantic text map could have potential in multiple applications, including scene understanding, navigation, augmented reality, and human-computer interaction.

\begin{table}[htbp]
	\setlength{\tabcolsep}{1mm}
	\caption{Results in an outdoor commercial center during the day. RPE (10 m) and APE (m)}
	\begin{center}
		\begin{tabular}{ccccccccc}
			\hline
			\multirow{2}{*}{\textbf{Seq.}} & \multicolumn{3}{c}{\textbf{ORB-SLAM}} & \multicolumn{2}{c}{\textbf{DSO}} &  \multicolumn{3}{c}{\textbf{TextSLAM}} \\ \cline{2-9} 
			&\multicolumn{1}{c}{LOOP} & \multicolumn{1}{c}{APE} & \multicolumn{1}{c}{RPE} & \multicolumn{1}{c}{APE} & \multicolumn{1}{c}{RPE} & \multicolumn{1}{c}{LOOP} & \multicolumn{1}{c}{APE} & \multicolumn{1}{c}{RPE} \\ \hline
			Outdoor\_1   &$\times$ & 1.159 & \bf{0.379} & 1.175 & 0.393 &$\surd$ & \bf{0.688} & 0.389 \\	
			Outdoor\_2   &$\times$& 1.340 & 0.511 & 1.280 & \bf{0.457} &$\surd$& \bf{0.561} & 0.470 \\	
			Outdoor\_3   &$\times$& 1.213 & 0.347 & 1.423 & 0.337 &$\surd$& \bf{0.807} & \bf{0.317} \\	
			Outdoor\_4   &$\surd$& \bf{0.116} & \bf{0.108} & 1.511 & 0.462 &$\surd$& 1.624 & 0.759 \\	
			Outdoor\_5   &$\surd$& \bf{0.175} & \bf{0.094} & 1.410 & 0.523 &$\surd$& 0.219 & 0.173 \\	
			Outdoor\_6   &$\times$& 1.491 & 0.462 & 1.450 & 0.299 &$\surd$& \bf{0.412} & \bf{0.238} \\	
			Outdoor\_7   &$\times$& 1.279 & 0.457 & 1.572 & 0.369 &$\surd$& \bf{0.563} & \bf{0.307} \\	
			Outdoor\_8   &$\times$& 1.358 & 0.529 & 1.642 & 0.620  &$\surd$& \bf{0.446} & \bf{0.249} \\	\hline 
		\end{tabular}
	\end{center}
	\emph{The tick '$\surd$' indicates a success loop closing and '$\times$' indicates no loop has been found. The smallest errors are in bold texts.}
	\label{tab:BiJiang_DayTraj}
\end{table}
	
	\begin{figure*}[!h]
		\centering  
		\includegraphics[width=\textwidth]{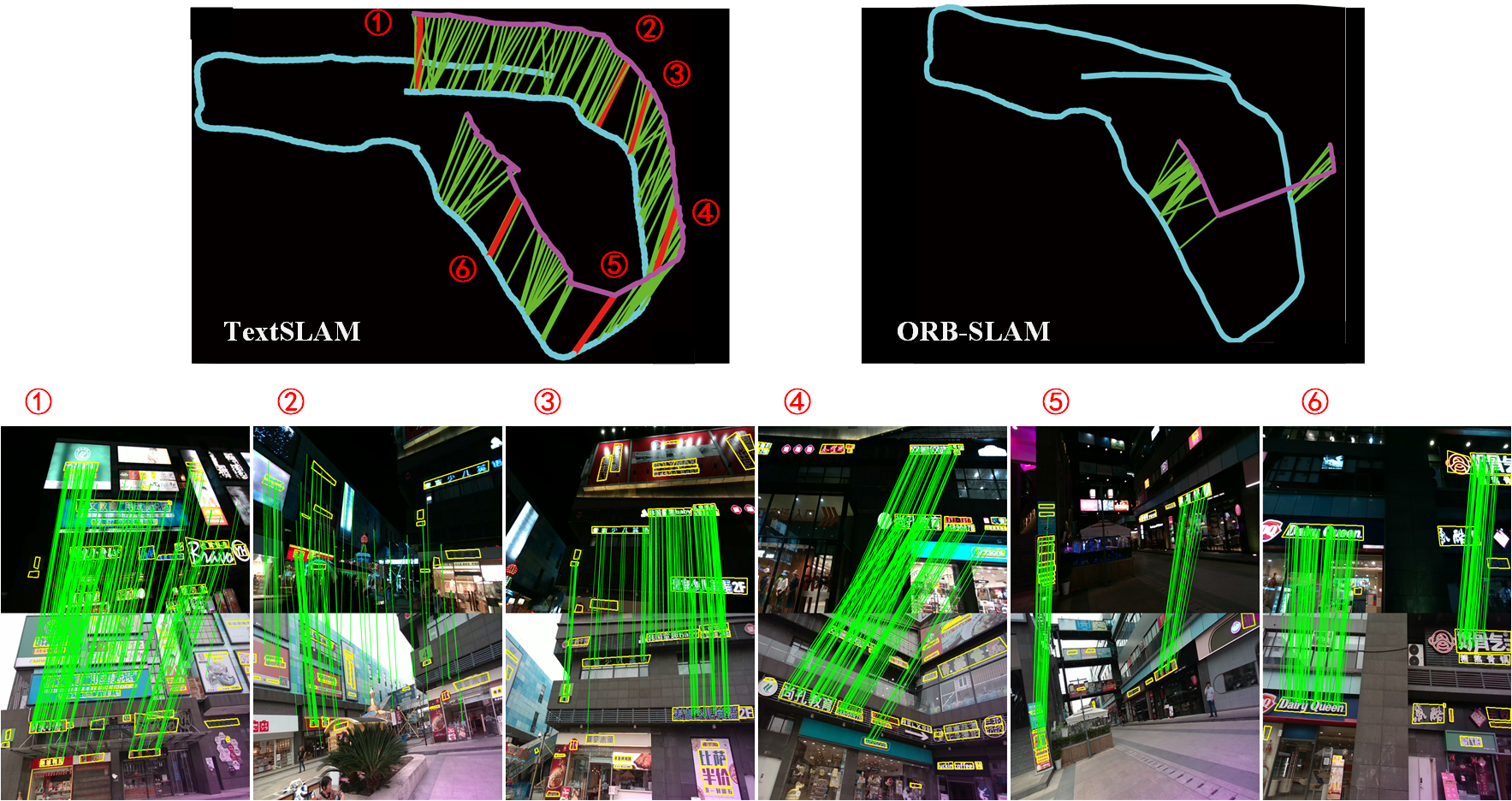}\\  
		\caption{ \textbf{Top row:} The blue trajectories are estimated by TextSLAM and ORB-SLAM respectively using day sequences, while the magenta trajectories are the localization results by registering the night images to the 3D map built during the day. We shift the night trajectories and connect the loop frames by green lines for better illustration. \textbf{Bottom row:} We also visualized \anno{the matched points} by TextSLAM in six different places. We can see that text-guided point matching correctly matched most of the text points despite large illumination changes.}
		\label{fig:BiJiang_Night_Traj}  
	\end{figure*}
	
\subsubsection{Day-night tests}

	\begin{figure*}[!h]
	\anno{
		\centering  
		\includegraphics[width=1\textwidth]{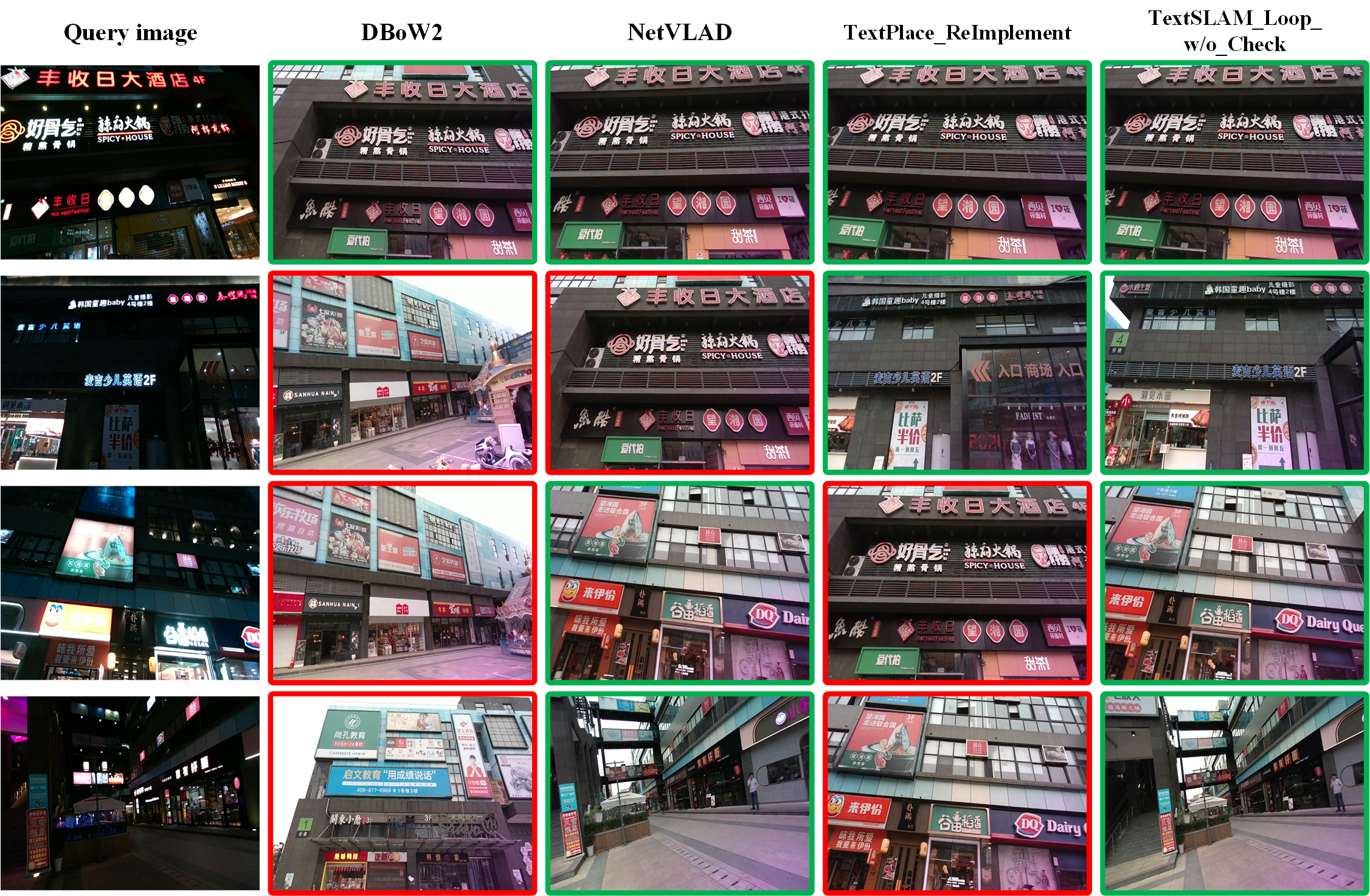}\\
		\caption{The image retrieval results of DBoW2, NetVLAD, TextPlace\_ReImplement and TextSLAM\_Loop\_w/o\_Check, respectively. The correct and wrong results are shown in green and red boxes, respectively.}
		\label{fig:localization_ImgRetrieval_examples}  	
	} 
	\end{figure*}
	
	\begin{figure*}[!h]
	\anno{
		\centering  
		\includegraphics[width=1\textwidth]{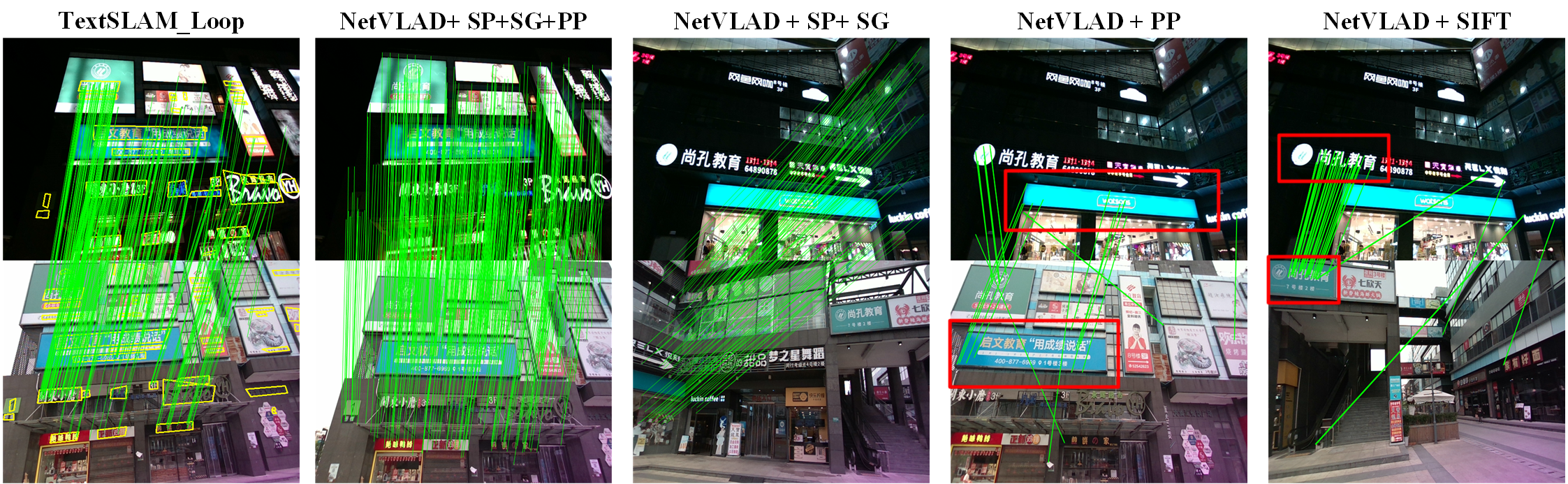}\\
		\caption{The point correspondence results of TextSLAM\_Loop, NetVLAD+SP+SG+PP, NetVLAD+SP+SG, NetVLAD+PP, and NetVLAD+SIFT, respectively. We can see that TextSLAM obtains correct point correspondence based on the invariant semantic text meaning. The deep-learning-based methods also work well during day-night change, as shown in the second and third columns. One failure case of the deep-learning-based method is the fourth column, where the similar blue boards (in red boxes) confuse the approach. SIFT also failed because of two similar boards (in red boxes) in different locations.}
		\label{fig:localization_PointMatch_examples} 
	}	 
	\end{figure*}

	Illumination change is one challenge that SLAM usually encounters. An extreme case is the day night variation. For example, we already built a map in the day, while we may want to reuse it at night. 
	To evaluate the performance towards this change, we collect night sequences in the same path as collecting those day sequences.
	
	\anno{To show how texts help matching scene points across day and night,} we implemented the localization-only version for both the TextSLAM \anno{(TextSLAM\_Loop)} and ORB-SLAM systems. 
	Based on the 3D model generated from the day sequence via each SLAM method, we test the localization performance using the night sequence.
	We present visual comparisons in \Fig{fig:BiJiang_Night_Traj}. The results show that our method can correctly locate a lot of frames, while ORB-SLAM works for only a few frames, implying the robustness of our method under such large illumination changes.
	
	\anno{We also compare our method with the state-of-the-art visual localization methods in the day-night tests. Because COLMAP failed to generate the ground truth trajectory of the night sequence, we follow the image retrieval evaluation protocol to use precision and recall for comparison.
	Additionally, the runtime is measured to show the efficiency of the approaches.
	Our day-night test contains 987 day keyframes as the database and 145 night keyframes as the queries.
	To obtain the ground truth of the night queries, we manually label each image pair among all 143115 ($987 \times 145$) pairs by checking if the two images belong to the same location.} 

	\anno{We compared the following methods from image retrieval, place recognition, and visual localization. Some of them are the top-ranked open source implementations (3rd, 4th, 34th ranked) in the Long-Term Visual Localization benchmark\cite{CTH2019LocalizationBenchmark}.}
	\anno{\begin{itemize}
		\item NetVLAD\cite{Arandjelovic2016CVPR}, the state-of-the-art deep learning-based image retrieval method. 
		\item DBoW2\cite{galvez2012bags}, which is widely used in existing visual SLAM systems\cite{mur2017orb}. 
		\item TextPlace\cite{hong2019textplace}, the place recognition method which uses 2D text extractions as the localization cue. Because TextPlace does not open its source code, we reimplement it for comparison, named as TextPlace\cite{hong2019textplace}\_ReImplement in results.
		\item NetVLAD\cite{Arandjelovic2016CVPR}+SuperPoint\cite{detone2018superpoint}+SuperGlue\cite{sarlin2020superglue} (abbr. NetVLAD+SP+SG). 3rd ranked method in \cite{CTH2019LocalizationBenchmark}.
		\item NetVLAD\cite{Arandjelovic2016CVPR}+SuperPoint\cite{detone2018superpoint}+SuperGlue\cite{sarlin2020superglue}+Patch2Pix\cite{zhou2021patch2pix} (abbr. NetVLAD+SP+SG+PP). 4th ranked method in \cite{CTH2019LocalizationBenchmark}.
		\item NetVLAD\cite{Arandjelovic2016CVPR}+Patch2Pix\cite{zhou2021patch2pix} (abbr. NetVLAD+PP). 34th ranked method in \cite{CTH2019LocalizationBenchmark}.
		\item NetVLAD\cite{Arandjelovic2016CVPR}+SIFT\cite{ng2003sift} (abbr. NetVLAD+SIFT). 
	\end{itemize}}

	\anno{Here, SIFT\cite{ng2003sift}, SuperPoint\cite{detone2018superpoint}, SuperGlue\cite{sarlin2020superglue}, and Patch2Pix\cite{zhou2021patch2pix} are used to refine the retrieval results from NetVLAD. SIFT\cite{ng2003sift} is a classic feature detector and descriptor. SuperPoint\cite{detone2018superpoint} is a learned interest point detector and descriptor. SuperGlue\cite{sarlin2020superglue} finds pointwise correspondences using a graph neural network with an attention mechanism. Patch2Pix\cite{zhou2021patch2pix} searches correspondences in a detect-to-refine manner (patch-level to pixel match).
	The results from TextSLAM localization-version are displayed as 'TextSLAM\_Loop'. We also present the results of TextSLAM localization-version without point matching and geometric validation as 'TextSLAM\_Loop\_w/o\_Check'.}
	
\begin{figure}[!h]
	\anno{
		\centering  
		\includegraphics[width=0.48\textwidth]{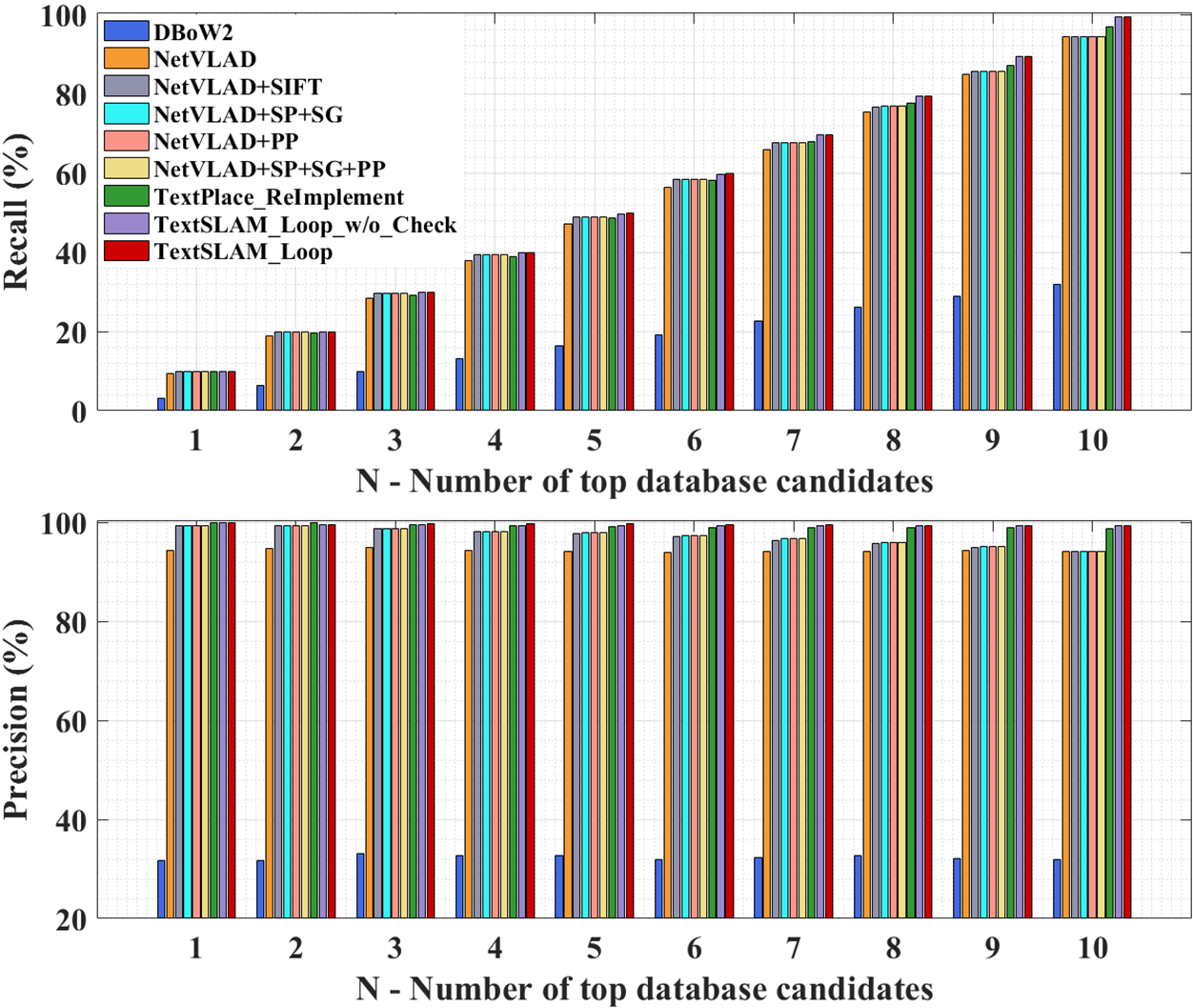}\\
		\caption{Recall and precision of the top-10 results. We normalize the recalls by dividing their maximum possible value: 10/\{average number of ground-truth pairs per query\} for better visualization.}
		\label{fig:localization}  
	}
	\end{figure}

	\anno{The deep learning-based methods ran on the NVIDIA RTX A6000 with the Intel i9-10980XE CPU of 128-GB RAM. Other methods ran in a single thread on an Intel Core i7- 9700K desktop computer with 32-GB RAM. For all compared methods, the top 10 candidates are retrieved and validated in further steps. For the methods that output point correspondence results, we adopt a geometric check (fundamental matrix estimation by RANSAC\cite{fischler1981random}) to reject outliers, and use the inlier number to rank the candidates from the previous step. The re-ranked results are taken as the final retrieval result. For the TextSLAM pipeline, we use PnP to reject the outliers. The precision and recall curves are displayed in \Fig{fig:localization}. To compare the efficiency of the approaches fairly, we randomly select 10 night queries to run each method repeatedly. The average 10-queries running time results are shown in \Tab{tab:exps_localization_time}. Several examples are shown in \Fig{fig:localization_ImgRetrieval_examples} and \Fig{fig:localization_PointMatch_examples}.}
	
	\anno{The results show that TextSLAM (with and without geometric validation) outperforms the state-of-the-art NetVLAD-based methods on this day-night test, while the latter was particularly trained to address significant illumination changes.  Interestingly, another text-based approach (Textplace\cite{hong2019textplace}) also performs better than NetVLAD-based methods. It implies that feature matching can greatly benefit from text semantics in text-rich environments. TextSLAM performs better than Textplace\cite{hong2019textplace} because Textplace\cite{hong2019textplace} uses 2D-2D text matching for image retrieval which is prone to false text detection and recognition. By contrast, TextSLAM matches the 2D text objects directly with the 3D text map where the semantics of 3D text objects are derived from multiple observations and hence are more reliable. The 2D-3D matching strategy also makes TextSLAM highly efficient as shown in \Tab{tab:exps_localization_time} because much fewer text objects in the 3D map are required to be matched.}	

	\begin{table}[htbp]
	\anno{
		\setlength{\tabcolsep}{1mm}
		\caption{Average runtimes of localization methods. (s)}
		\begin{center}
			\scriptsize
			\setlength{\tabcolsep}{0.5mm}
			\begin{tabular}{lccc}
				\hline
				\multicolumn{1}{c}{{\bf Methods}} &\multicolumn{1}{c}{{\bf Image retrieval}} & \multicolumn{1}{c}{\bf {Point matching}} & \multicolumn{1}{c}{{\bf Geometric check}} \\ \hline 
				NetVLAD\cite{Arandjelovic2016CVPR}  & $ 0.055 \pm 0.001 $ & -- & -- \\
				NetVLAD+PP & $0.055 \pm 0.001 $ & $ 1.427 \pm0.157$ & $ 7.999\pm 0.508$ \\ 	
				NetVLAD+SP+SG & $0.055 \pm 0.001 $ & $1.625 \pm0.034 $ & $ 5.992\pm0.683$ \\	
				NetVLAD+SP+SG+PP  & $ 0.055 \pm 0.001 $ & $ 2.183 \pm0.168$ & $ 6.332\pm 0.674$ \\	
				TextPlace\cite{hong2019textplace}\_ReImplement   & $ 0.358 \pm 0.077 $ & -- & -- \\
				TextSLAM\_Loop  & $\bm{0.005} \pm \bm{0.002}$ &  $ \bm{0.037}\pm \bm{0.011} $ & $\bm{0.083} \pm \bm{0.018}$ \\
				\hline
			\end{tabular}
		\end{center}
		\emph{The abbreviations 'PP', 'SP', 'SG' represent Patch2Pix\cite{zhou2021patch2pix}, SuperPoint\cite{detone2018superpoint}, and SuperGlue\cite{sarlin2020superglue}, respectively.}
		\label{tab:exps_localization_time}
	}
	\end{table}

\anno{\subsection{Runtime Analysis}}

	\anno{We selected four typical sequences from above experiments to test the runtime of TextSLAM. We ran TextSLAM in a single thread on an Intel Core i7-9700K desktop computer with 32-GB RAM. The text extractor\cite{zhang2019feasible} using a neural network ran on the NVIDIA GeForce RTX 2080 Ti. The runtimes of major components in the system are present in \Tab{tab:exps_time}. We can see that text extraction requires much more time than other front-end components (point extraction and pose estimation). It is therefore the bottleneck of TextSLAM’s efficiency. But this problem can be solved if a highly efficient text extractor appears.}
	
	\anno{We also present the running time of our text-based loop closing in \Tab{tab:exps_loop_time}, which includes the average single-threaded runtime for loop detection (Section 4.5.1), relative transformations calculation (Section 4.5.2), and loop correction.}
	
	\begin{table}[htbp]
	\anno{
		\scriptsize
		\setlength{\tabcolsep}{0.1mm}
		\caption{Runtime analysis of our method. (s)}
		\begin{center}
			\begin{tabular}{ccccc}
				\hline
				\multicolumn{1}{c}{\textbf{Seq.($\bm{\#}${\bf Text}) }} &\multicolumn{1}{c}{{\bf Text extraction}} &\multicolumn{1}{c}{{\bf Point extraction}} & \multicolumn{1}{c}{{\bf Pose estimation}} & \multicolumn{1}{c}{{\bf LocalBA}}  \\ \hline	
				LIndoorLoop\_07 (2)  & $ 0.227 \pm 0.037 $ & $ 0.008\pm 0.002 $ & $ 0.004 \pm 0.003 $ & $ 0.489 \pm 0.142  $ \\ 	
				LIndoorLoop\_01 (3)  & $ 0.245 \pm 0.048 $ & $0.008 \pm 0.002$ & $ 0.008 \pm 0.009 $ & $ 0.660 \pm 0.366 $\\ 	
				Indoor\_07 (22)  & $ 0.573 \pm 0.106 $ & $ 0.008 \pm 0.002 $ & $ 0.064 \pm 0.019 $ & $ 2.788 \pm 0.691 $\\ 	
				Outdoor\_1 (25)  & $ 0.547 \pm 0.195 $ & $ 0.010 \pm 0.002$ & $ 0.068 \pm 0.042 $ & $ 2.438 \pm 1.226 $\\ 	
				\hline
			\end{tabular}
		\end{center}
		\emph{
			\textbf{$\bm{\#}$Text} in the first column means the average observed text object number per frame. \\ 
			The second column \textbf{Text extraction} represents the runtime per frame, including the text detection, recognition as well as text representative pixel extraction.}
		\label{tab:exps_time}
	}
	\end{table}
	
	\begin{table}[htbp]
	\anno{
			\scriptsize
			\setlength{\tabcolsep}{1.5mm}
			\caption{Runtime analysis of the loop of our method.}
			\begin{center}
				\begin{tabular}{cccc}
					\hline
					\multicolumn{1}{c}{\textbf{Seq.}} &\multicolumn{1}{c}{{\bf Loop detection}} & \multicolumn{1}{c}{{\bf Sim3 calculation}} & \multicolumn{1}{c}{{\bf Loop correction}} \\ \hline 
					LIndoorLoop\_07  & 0.002s & 0.095s & 23.761s \\
					LIndoorLoop\_01$^1$ & 0.009s & 0.195s & 33.626s \\
					LIndoorLoop\_01$^2$ & 0.008s & 0.127s & 64.380s \\
					Outdoor\_1   & 0.008s & 0.360s & 85.855s \\
					\hline
				\end{tabular}
			\end{center}
			\emph{The superscript $^1$ and $^2$ indicate the first loop and the second loop of LIndoorLoop\_01, respectively.}
			\label{tab:exps_loop_time}
		}
	\end{table}

\section{Limitation}

		Here, we discuss the limitations of our method.
		Firstly, TextSLAM relies on the text objects in the scene. When no text object exists, TextSLAM will switch into the point-only mode. So no high-level information can be used for improving SLAM in this case. 
		Fortunately, \anno{many daily scenes, such as city commercial plazas or shopping malls,} are full of text objects. 
		\anno{Secondly, TextSLAM still relies on the texture in the environment and easily fails in the scene with little textures, similar to other visual SLAM methods.}	
		\anno{Thirdly,} the text detector still does not work perfectly. False detection and recognition still happen frequently. For example, some windows and bench legs are recognized as text strings. Moreover, these text extraction methods require lots of training data that needs great effort for labeling. 
		Finally, our TextSLAM cannot run in real-time currently because of the time-consuming operations on text extraction and the back-end optimization. Nevertheless, the system can be further optimized for efficiency.
		
		

\section{Conclusion}
	In this paper, we fully explore the text objects both geometrically and semantically and propose a novel visual SLAM approach tightly coupled with the semantic text features, where text objects are regarded as local planar patches with rich textual and semantic meaning.
	\anno{We tested our method in various indoor and outdoor environments} involving challenges such as fast camera motions, viewpoint changes, and day-night illumination shifts. \anno{The results show that with the help of scene texts, our method outperforms the state-of-the-art methods including SLAM and visual localization in terms of accuracy and robustness}, indicating the benefits of integrating semantic text objects into visual SLAM. The 3D text map produced by our system can serve as an important medium to bridge humans and robots. We hope our work could inspire more explorations to the semantic texts in various applications in robotics, navigation, human-computer interaction, AR and VR, etc. 

\ifCLASSOPTIONcompsoc
  \section*{Acknowledgments}
\else
  \section*{Acknowledgment}
\fi

The authors would like to thank Shanghai Alpha square for the support of data collection.

\ifCLASSOPTIONcaptionsoff
  \newpage
\fi



\bibliographystyle{IEEEtran}
\bibliography{IEEEabrv,ReferenceTex}
%
%
%

%

\begin{IEEEbiography}[{\includegraphics[width=1in,height=1.25in,clip,keepaspectratio]{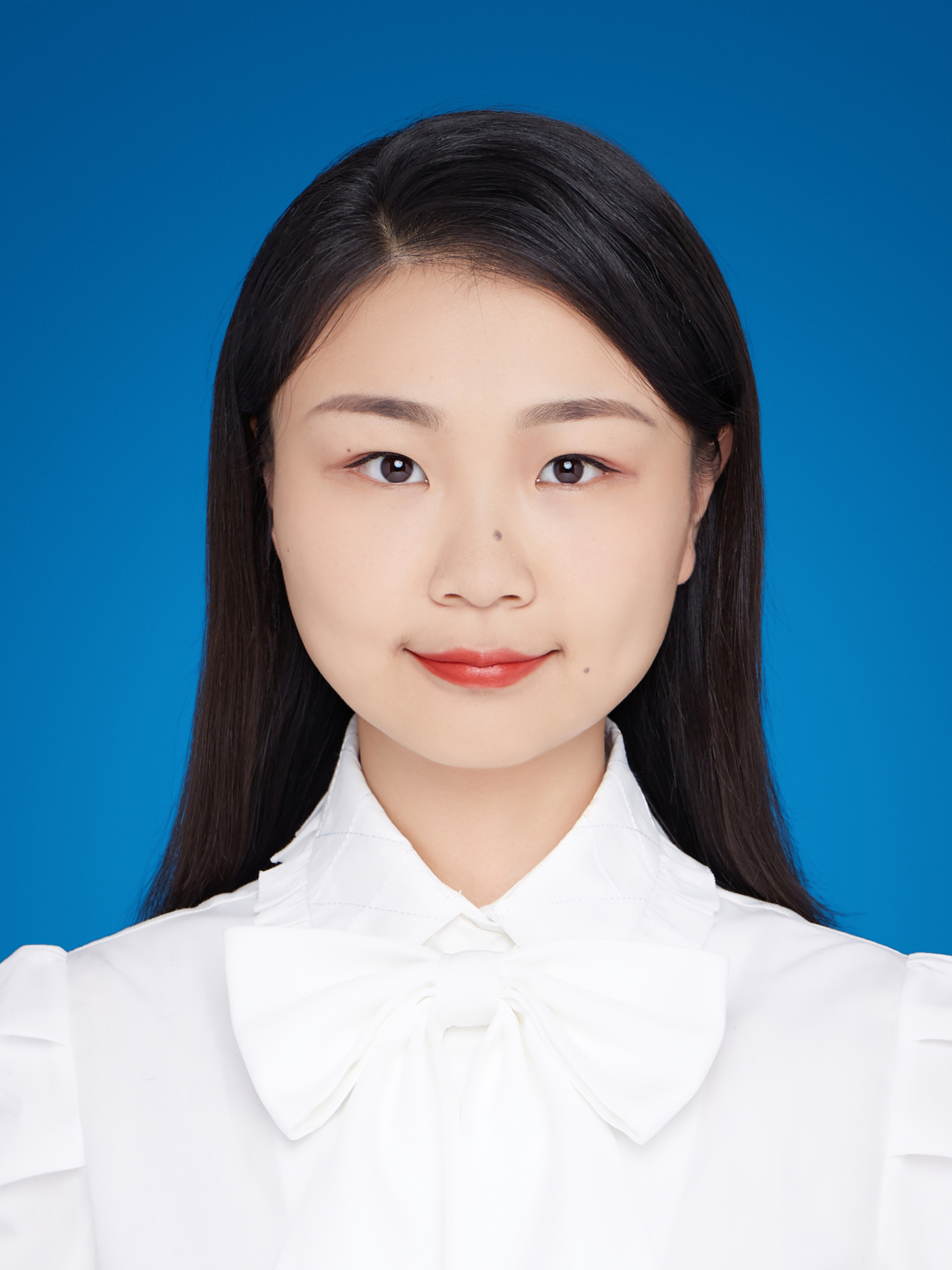}}]{Boying Li}
	received the B.Sc. degree in automation engineering from Northwestern Polytechnical University, Xi’an, China, in 2016. She is currently pursuing the Ph.D. degree with the Department of Electronic Engineering, Shanghai Jiao Tong University, Shanghai, China. Her research interests include 3D vision and visual SLAM.
\end{IEEEbiography}

\begin{IEEEbiography}[{\includegraphics[width=1in,height=1.25in,clip,keepaspectratio]{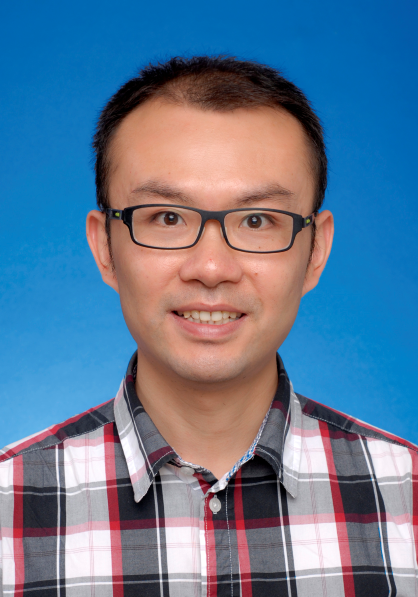}}]{Danping Zou}
	is an Associate Professor in School of Electronic Information and Electrical Engineering, Shanghai Jiao Tong University. He is now leading the Vision and Intelligence SYStem (ViSYS) group in Institute for Sensing and Navigation. His research mainly focuses on 3D vision and autonomous systems. 
\end{IEEEbiography}

\begin{IEEEbiography}[{\includegraphics[width=1in,height=1.25in,clip,keepaspectratio]{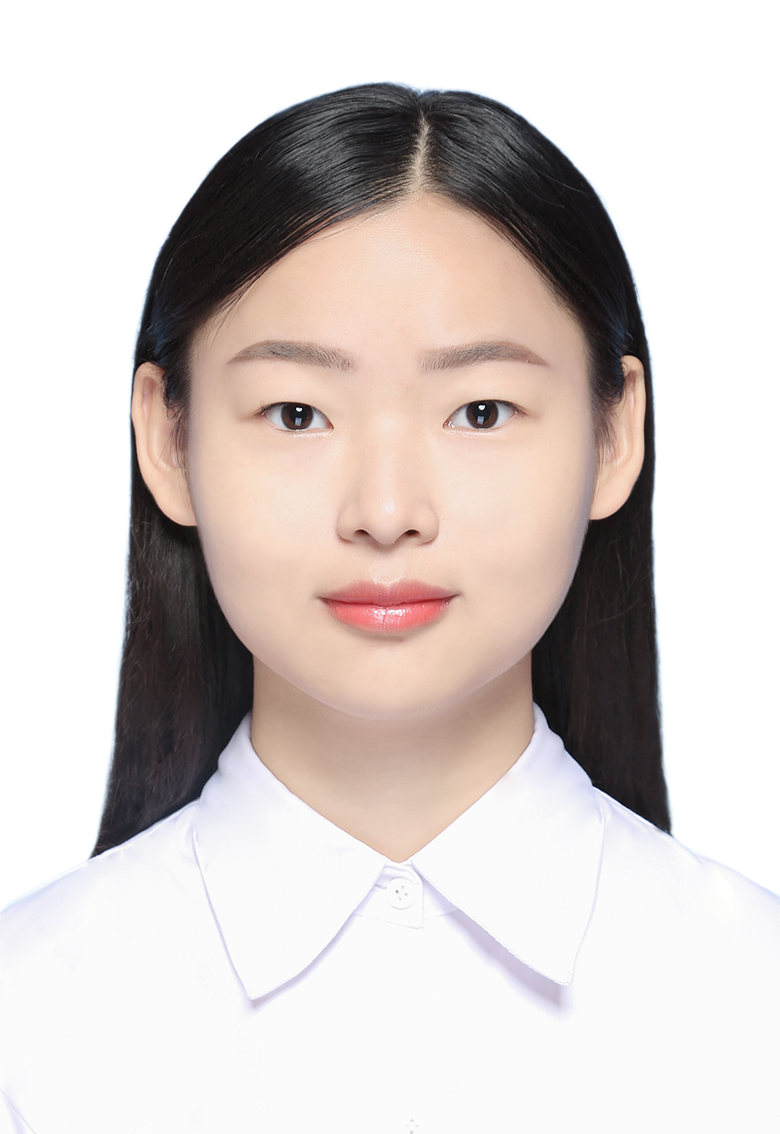}}]{Yuan Huang}
	received the B.Sc. degree in communication engineering from Huazhong University of Science and Technology, Wuhan, China, in 2019. She is currently pursuing the M.Eng. degree with the Department of Electronic Engineering, Shanghai Jiao Tong University, Shanghai, China.
	Her research interests include computer vision and structure-from-motion. 
\end{IEEEbiography}

\begin{IEEEbiography}[{\includegraphics[width=1in,height=1.25in,clip,keepaspectratio]{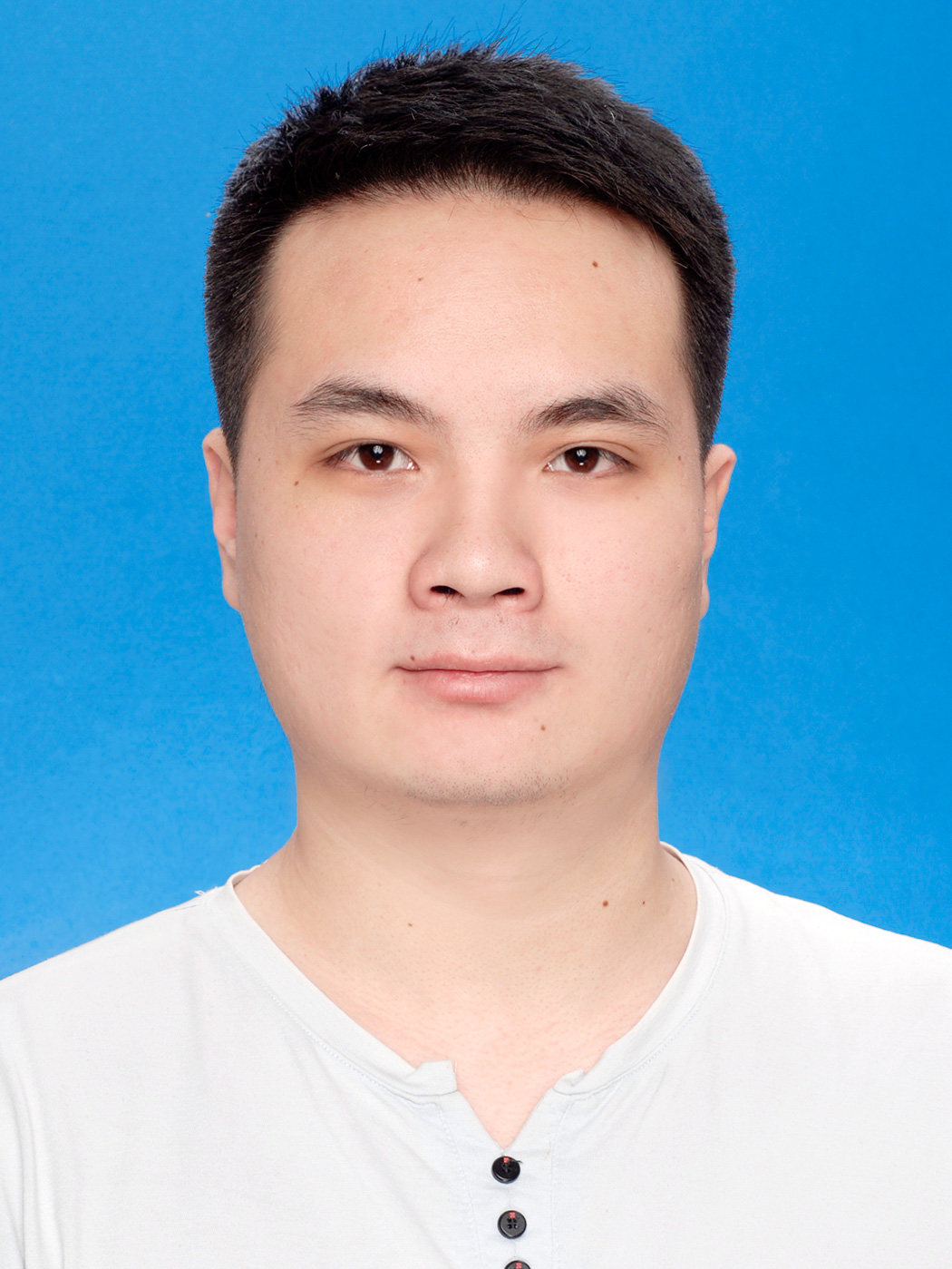}}]{Xinghan Niu}
	received the B.E degree in information engineering from Shanghai Jiao Tong University, Shanghai, China, in 2020. He is currently pursuing the master degree with the Department of Electronic Engineering, Shanghai Jiao Tong University, Shanghai, China.
	His research interests include depth map prediction and machine learning.
\end{IEEEbiography}

\begin{IEEEbiography}[{\includegraphics[width=1in,height=1.25in,clip,keepaspectratio]{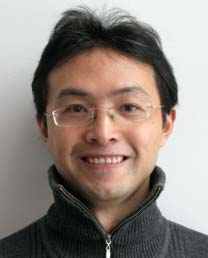}}]{Ling Pei}
	received the Ph.D. degree from Southeast University, Nanjing, China, in 2007. From 2007 to 2013, he was a Specialist Research Scientist with the Finnish Geospatial Research Institute, Masala, Finland. He is currently a Professor with the School of Electronic Information and Electrical Engineering, Shanghai Jiao Tong University, Shanghai, China. He has authored or coauthored over 90 scientific articles. He is an inventor of 24 patents and pending patents. His main research is in the areas of indoor/outdoor seamless positioning, ubiquitous computing, wireless positioning, bio-inspired navigation, context-aware applications, location-based services, and navigation of unmanned systems.
	Dr. Pei was a recipient of Shanghai Pujiang Talent in 2014.
\end{IEEEbiography}

\begin{IEEEbiography}[{\includegraphics[width=1in,height=1.25in,clip,keepaspectratio]{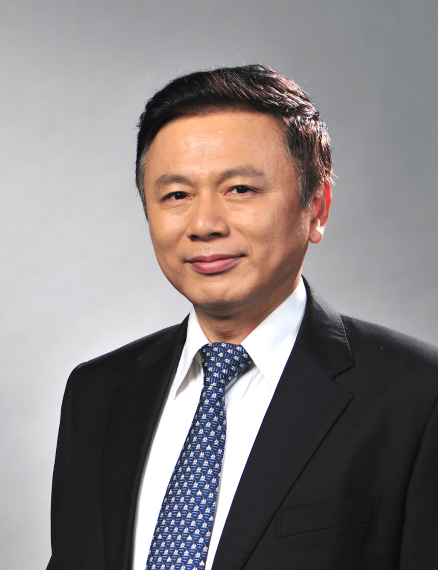}}]{Wenxian Yu}
	received the B.S., M.S., and Ph.D. degrees from the National University of Defense Technology, Changsha, China, in 1985, 1988, and 1993, respectively. From 1996 to 2008, he was a Professor with the College of Electronic Science and Engineering, National University of Defense Technology, where he served as the Deputy Head of the College and the Assistant Director of the National Key Laboratory of Automatic Target Recognition. From 2009 to 2011, he was the Executive Dean of the School of Electronic, Information, and Electrical Engineering, Shanghai Jiao Tong University, Shanghai, China. He is currently a Yangtze River Scholar Distinguished Professor and the Head of the research part at the School of Electronic, Information, and Electrical Engineering, Shanghai Jiao Tong University. His research interests include remote sensing information processing, automatic target recognition, and multisensor data fusion. He has published over 200 research papers in these areas. 
\end{IEEEbiography}

%
%
%




\end{document}